%% file: main.tex
\documentclass[journal]{IEEEtran}

\usepackage{amsmath,amssymb,amsfonts}
\usepackage{algorithm}
\usepackage{algorithmic}
\usepackage{array}
\usepackage{bbm}
\usepackage{booktabs}
\usepackage{cite}
\usepackage{graphicx}
\usepackage{multirow}
\usepackage[caption=false,font=footnotesize]{subfig}
\usepackage{textcomp}
\usepackage{url}
\usepackage{xcolor}

\definecolor{myBrightRed}{rgb}{1, 0, 0}
\definecolor{myBrightGreen}{rgb}{0, 1, 0}
\definecolor{myBrightBlue}{rgb}{0, 0, 1}
\definecolor{myYellow}{rgb}{1, 1, 0}
\definecolor{myMagenta}{rgb}{1, 0, 1}
\definecolor{myCyan}{rgb}{0, 1, 1}
\definecolor{myDarkRed}{rgb}{0.5, 0, 0}
\definecolor{myDarkGreen}{rgb}{ 0.0, 0.5, 0.0}
\definecolor{myDarkBlue}{rgb}{0.0, 0.0, 0.5}
\definecolor{myOrange}{rgb}{1.0, 0.65, 0.0}
\definecolor{myPurple}{rgb}{0.54, 0.17, 0.89}
\definecolor{myGray}{rgb}{0.5, 0.5, 0.5}
\definecolor{myOlive}{rgb}{0.5, 0.5, 0.0}
\definecolor{myTeal}{rgb}{0.0, 0.5, 0.5}
\definecolor{myCrimson}{rgb}{0.86, 0.08, 0.24}
\definecolor{myTurquoise}{rgb}{0.13, 0.7, 0.67}

\begin{document}

\IEEEpubid{This work has been submitted to the IEEE for possible publication. Copyright may be transferred without notice, after which this version may no longer be accessible.}

\title{Scalable Hypergraph Structure Learning with\\ Diverse Smoothness Priors}

\author{Benjamin~T.~Brown,
        Haoxiang~Zhang,
        Daniel~L.~Lau,~\IEEEmembership{Senior~Member,~IEEE}
        and~Gonzalo~R.~Arce,~\IEEEmembership{Life~Fellow,~IEEE}

\thanks{
B.~T.~Brown, H.~Zhang, and D.~L.~Lau are with the Department of Electrical and Computer Engineering, University of Kentucky, Lexington, KY 40506, USA.\\
\indent G.~R.~Arce is with the Department of Electrical and Computer Engineering, University of Delaware, Newark, DE 19716, USA.\\
\indent This work was partially supported by the National Science Foundation under grants 1815992 and 1816003 and the AFOSR award FA9550-22-1-0362.}}

\maketitle

\input{Sections/abstract}
\input{Sections/introduction}

\input{Sections/preliminaries}
\input{Sections/HGSI}
\input{Sections/HSLS}
\input{Sections/experiments}
\input{Sections/conclusion}
\input{Sections/acknowledgement}
\input{Sections/appendix}

\bibliographystyle{IEEEtran}
\bibliography{references}

\end{document}

%% file: Sections/abstract.tex
\begin{abstract} \label{sec:Abstract}
In graph signal processing, learning the weighted connections between nodes from a set of sample signals is a fundamental task when the underlying relationships are not known a priori. This task is typically addressed by finding a graph Laplacian on which the observed signals are smooth. With the extension of graphs to hypergraphs -- where edges can connect more than two nodes -- graph learning methods have similarly been generalized to hypergraphs. However, the absence of a unified framework for calculating total variation has led to divergent definitions of smoothness and, consequently, differing approaches to hyperedge recovery. We confront this challenge through generalization of several previously proposed hypergraph total variations, subsequently allowing ease of substitution into a vector based optimization. To this end, we propose a novel hypergraph learning method that recovers a hypergraph topology from time-series signals based on a smoothness prior. Our approach, designated as Hypergraph Structure Learning with Smoothness (HSLS), addresses key limitations in prior works, such as hyperedge selection and convergence issues, by formulating the problem as a convex optimization solved via a forward-backward-forward algorithm, ensuring guaranteed convergence. Additionally, we introduce a process that simultaneously limits the span of the hyperedge search and maintains a valid hyperedge selection set. In doing so, our method becomes scalable in increasingly complex network structures. The experimental results demonstrate improved performance, in terms of accuracy, over other state-of-the-art hypergraph inference methods; furthermore, we empirically show our method to be robust to total variation terms, biased towards global smoothness, and scalable to larger hypergraphs.
\end{abstract}

%% file: Sections/introduction.tex
\section{Introduction} \label{sec:Introduction}

\begin{figure}[!t]
    \centering
    \includegraphics[width=0.8\linewidth]{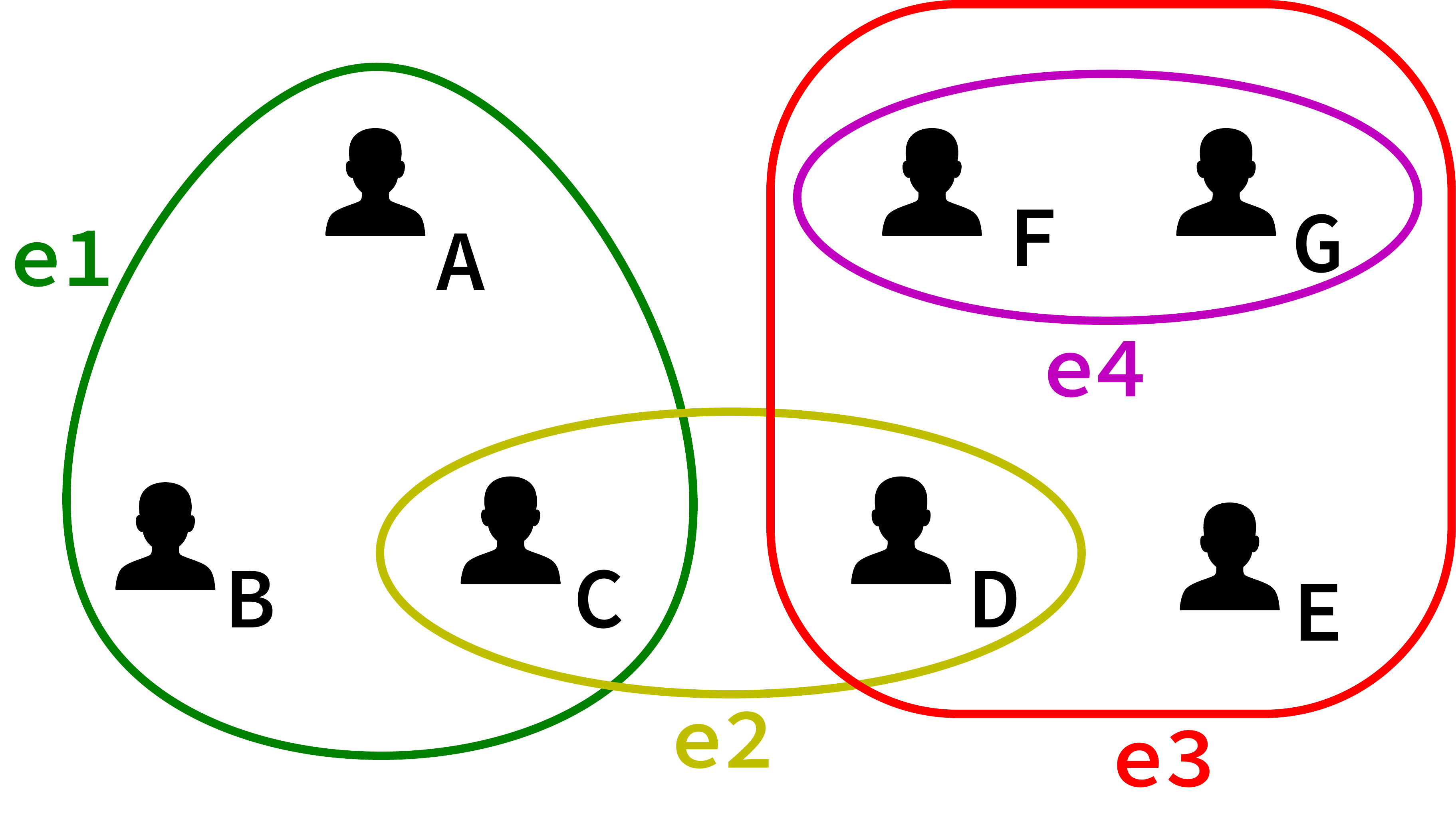}
    \caption{Example of a hypergraph network capturing higher order relationships. Nodes \textbf{A} through \textbf{F} are workers, and the colored connections \textbf{e1} through \textbf{e4} are hyperedges relating workers who are on the same project together.}
    \label{fig:hypergraph}
\end{figure}

\IEEEPARstart{H}{ypergraphs} are considered as generalized graphs that capture higher order relationships~\cite{HypergraphIntro}. While graphs encode pairwise relationships between nodes through edges, the higher order nature of hypergraphs extends node relations to allow an arbitrary number of nodes to be connected by a hyperedge. Figure~\ref{fig:hypergraph} contains a sample hypergraph displaying these higher order connections where nodes are considered workers and hyperedges connect workers who are collaborating on a project. Hypergraph models have been applied in the contexts of email networks~\cite{Email}, multi-cell multi-user power allocation~\cite{Power}, machinery fault diagnosis~\cite{Fault}, short text classification~\cite{Text}, traffic forecasting~\cite{Traffic}, and brain connectivity networks~\cite{Brain}. In the application of brain connectivity, nodes are regions of the brain specified by lobe and hemisphere, and hyperedges represent frequency bands. Any number of nodes could be connected in a hyperedge if, within the frequency band, there was an appropriate measurement of event-related coherence based on MEG measurements. Hypergraphs can also be used in scenarios where graphs already exist, but the relations could be extended to higher order. Examples of this lie in 3-D point cloud processes~\cite{Image}, object classification~\cite{Object}, pixel segmentation~\cite{Segmentation}, and noise reduction~\cite{Noise}.

\IEEEpubidadjcol

While there are situations in which the choice of a network structure is naturally dependent on the application, it is not always a known quantity. Furthermore, even if a prior structure is established, it may not be optimized to the nodal relations. In such cases, the structure of the network could be recovered via learning methods. This structure recovery is of common interest with graphs and is known as graph learning or graph structure inference. Such applications of graph learning include connectome networks~\cite{BrainNetworks}, social networks~\cite{SocialNetworks}, and diffusion including atmospheric tracer and mobility networks~\cite{GraphDiffusion}. Hypergraphs can be used to generalize graph learning into hypergraph learning, or hypergraph structure inference, which involves the recovery of a hypergraph structure from a set of observations, where the observations are hypergraph signals.

Some of the more prevalent and successful learning methods on traditional graphs are based on maximizing the smoothness of a set of sample signals across the graph structure where smoothness is the property describing the way a signal travels between connected nodes. At the core of recovering the topology is the concept of total variation (TV) -- the measurement of this smoothness property or, more precisely, a measure of how not smooth a particular signal is where maximizing smoothness is synonymous with minimizing the signal's total variation. To this end, Dong et al.~\cite{Dong} use a factor analysis model with a Gaussian probabilistic prior to introduce a method of recovering a graph Laplacian matrix that is effective on smooth graph signals.

\begin{figure}[!t]
    \centering
    \includegraphics[width=0.75\linewidth]{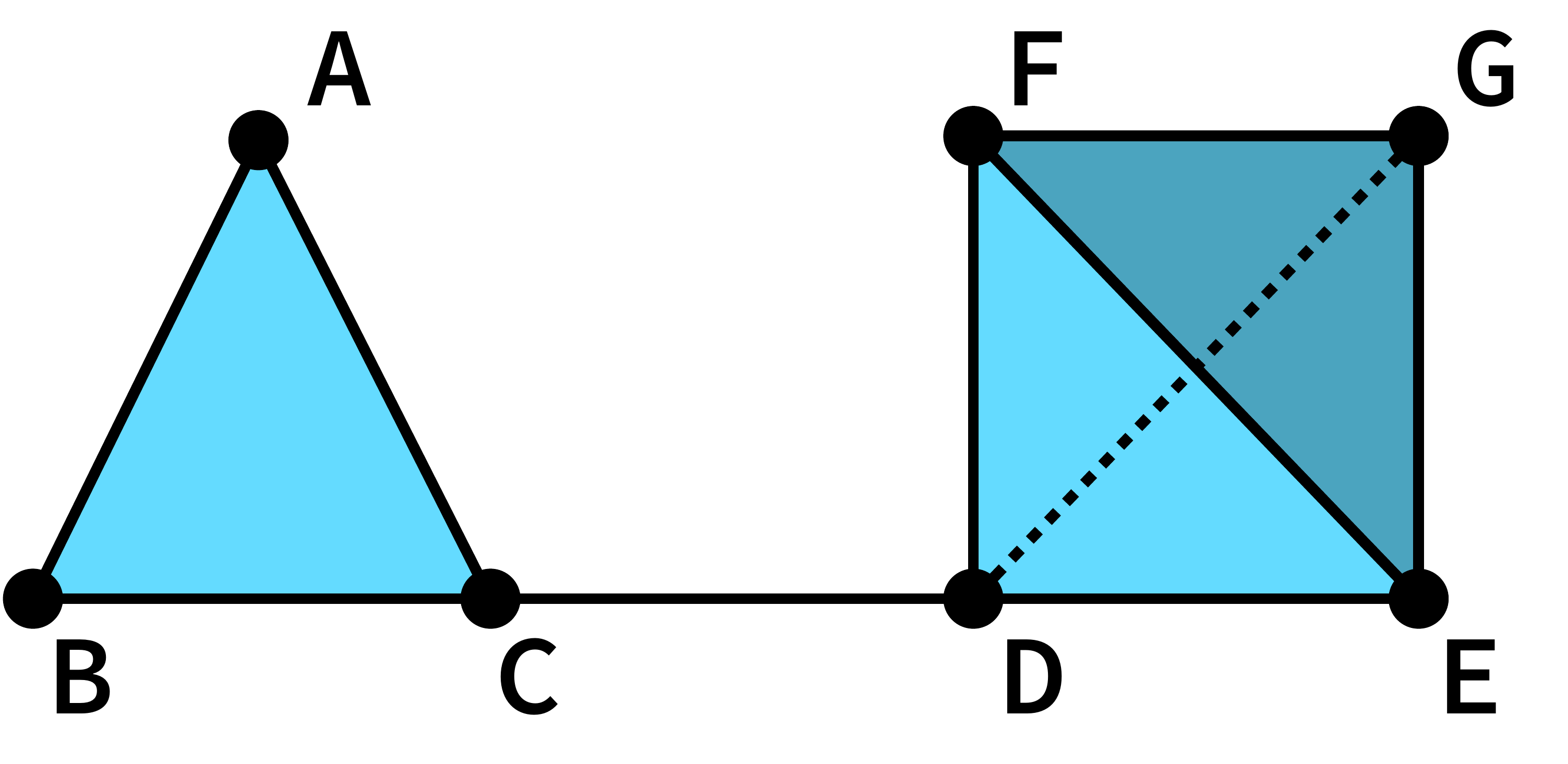}
    \caption{Example of a simplicial complex representing a similar higher order network as in Fig.~\ref{fig:hypergraph} with nodes being workers. The key difference here is that Fig.~\ref{fig:hypergraph} has a structure of four hyperedges while this network has a 0-simplex for every node, a 1-simplex for every line (dashed or solid lines), a 2-simplex for every triangle (colored faces), and a 3-simplex for every tetrahedron (groups of 4 nodes). Each hyperedge has been converted into its highest order simplex containing all lower order simplices.}
    \label{fig:simplex}
\end{figure}

With inspiration from Dong et al., Kalofolias~\cite{Kalofolias} developed a graph learning method based on smooth signals that infers a graph weighted adjacency matrix. Both of these methods revolve around maximizing graph signal smoothness, or equivalently, minimizing graph total variation. These works are the basis for several other graph learning related literature~\cite{DongKaloCite1,DongKaloCite2,DongKaloCite3}. Furthermore, the smoothness prior is effective and abundantly used in diverse graph learning contexts~\cite{MoreGraphSmoothness,MoreGraphSmoothness2,MoreGraphSmoothness3,MoreSmoothness4}. Reiterating that hypergraphs can be viewed as generalized graphs, it is natural that establishing a smoothness prior on hypergraph signals would be beneficial for hypergraph structure inference. However, while total variation on graphs have a well established connection to the notion of frequency in the graph domain, with smaller total variation indicating a lower frequency signal, the spectral relationship between hypergraphs and frequency via total variation is still an area of emerging research~\cite{HypergraphFreq,T-HGSP,Karelia,Fuli}.

Some existing hypergraph learning works employ the use of hypergraph structure inference for downstream tasks. Gao et al.~\cite{HGNN+} developed a hypergraph convolution based neural network for representation learning known as HGNN+. This method uses a two step learning process where the first is modeling the hypergraph structure and the second is hypergraph convolution, which can be thought of as part of the downstream task. Gao et al. offers several hyperedge group generation options, but one in particular makes use of node features. In the node feature space, each node's K-nearest neighbors (K-NN) are selected and formed into a hyperedge with the original node. Alternatively, the K-NN selection can be based on some maximum feature distance. In a separate example, Yang et al.~\cite{Volterra} implement the Volterra model to estimate higher-order connections through use of structural equation models of nonlinear relationships. This learned structure is subsequently used in the task of link prediction for social networks.

In another work, Duta and Liò~\cite{SPHINX} develop a hypergraph learning method known as SPHINX that is based on the use of multi-layer perceptrons (MLP) to iteratively produce an incidence matrix with entries being probabilities that nodes belong to hyperedges. From this, a sampling method leverages the probabilities to group nodes into hyperedges. Similarly, Bollengier et al.~\cite{Bollengier} use a deep embedded clustering method to generate the same kind of probabilistic incidence matrix for the purpose of hypergraph structure inference. All of these methods can be used with node features to learn a hypergraph structure. However, none of them make use of signal smoothness to refine their hyperedge selection.

Zhang et al.~\cite{Zhang} propose dynamic hypergraph learning that alternates between updating the hypergraph incidence matrix and a label projection matrix. The problem formulation makes use of a matrix based smoothness prior on both the label projections and the node features. Another smoothness based method from Tang et al.~\cite{DualSmoothness} performs community detection on a line graph to form hyperedges. They use weighted squared Euclidean distance as total variation on a graph before proceeding to line graph construction and subsequent hypergraph structure recovery. While the methods of Zhang et al. and Tang et al. identify node signal smoothness to be of importance, the lack of uniformity of the smoothness measurement discourages judgments between which one is more effective, or if it is application dependent.

Simplicial complexes are hierarchical structures that are an alternative for higher order network representation. Similar to hypergraphs, simplicial complexes can group together any number of nodes in a $c$-simplex where $c$ is one less than the number of nodes in the simplex, representing its geometric dimension. However, simplicial complexes diverge from hypergraph structures in that any $c$-simplex must also include all subset simplices from dimension $c-1$ to $0$~\cite{Simplex1}. This inherently limits the representational ability of simplicial complexes to densely connected spaces where all $c$ dimensions must be considered. Hypergraphs do not share this property, and are free to group any number of nodes in any subset without limitation. Figures~\ref{fig:hypergraph} and~\ref{fig:simplex} illustrate this comparison where the simplicial complex represents the same highest order connections as the hypergraph, but must also include all unique subsets in the topology. Furthermore, a hypergraph could represent a simplicial complex, and therefore is the more robust choice of network structure, at the sacrifice of scalability. So, while simplicial complexes are valid higher order structures that have been inferred from nodal signals~\cite{Simplex2,Simplex3,Simplex4}, this work will focus on hypergraph recovery since hypergraphs are inherently different structures~\cite{Simplex1} which have a broader range of applications.

Now as with graphs, hypergraph smoothness and total variation can be referenced interchangeably with total variation being the measurement of smoothness. However, there is disagreement within the hypergraph signal processing literature regarding the exact form of the total variation cost function. This diversity of smoothness measures can be attributed to the lack of a unified hypergraph signal processing framework. Consequently, there has been little work done to investigate the effects of different hypergraph total variations in structure recovery. In doing such research, this would provide insight into the validity of proposed hypergraph smoothness definitions. Case in point, Tang et al.~\cite{Tang} propose the hypergraph learning method Hypergraph Structure Inference Under Smoothness Prior (HGSI). The core of HGSI is a minimization problem that generates a vector-based solution capturing the hyperedges. Each term in the objective function controls a different aspect of the resultant hypergraph structure. Prominently discussed is the use of a smoothness prior through a total variation term limited to the metric equivalent of a sum of squared maximum differences between pairwise nodes derived by expansion of the hyperedges.

Following their outline, we propose a novel inference method known as Hypergraph Structure Learning with Smoothness (HSLS). Our method updates and improves upon the approach of HGSI in several ways. We allow our smoothness term to accommodate an assortment of total variation definitions taking into account linear summations and nonlinear maximums of squared and absolute differences between nodes. We also update the degree positivity term to operate on a dual variable. In doing so, the closed form optimal solution of HGSI is avoided and replaced with a primal-dual algorithmic approach, which benefits in several ways including more equitable hyperedge selection. Scalar parameters are introduced to enhance or diminish the effects of certain terms such as the degree of nodes. Finally, through an investigation of the use of K-NN to reduce the number of hyperedge possibilities, we propose a method that is scalable without compromising the span of the hyperedge search. Subsequently, we avoid limiting the resulting learned structure which followed assumptions that did not always maintain a desirable set of hyperedges. Experimentally, we show our method, across the majority of scenarios and metrics, outperforms HGSI, the K-NN approach of Gao et al., and a tensor based method from Pena-Pena~\cite{Karelia} known as PDL-HGSP. We also demonstrate our method's ability to prioritize smooth structure recovery and improve upon a pre-existing learning application.

Our work addresses limitations of prior research with the following novel contributions:
\begin{enumerate}
    \item We formulate a hypergraph learning model HSLS based on a smoothness prior which improves upon existing hypergraph learning works.
    \item HSLS is structured such that a variety of smoothness, or total variation, terms can be substituted, such that we can test the efficacy of hypergraph smoothness definitions in varying applications.
    \item We demonstrate why certain assumptions in previous works are undesirable and design HSLS to specifically avoid these issues as they relate to hyperedge selection and solution convergence. In doing so, our learning method becomes scalable to larger problems and adheres to a more complete view of the span of hyperedge possibilities.
    \item We generate and analyze original experimental results using HSLS in three application domains, each providing new empirical results both for the proposed method and smoothness definitions.
\end{enumerate}
The rest of the paper is organized as follows: Section~\ref{sec:Preliminaries} covers preliminary information including hypergraph notation, optimization, and total variation measures. Section~\ref{sec:HGSI} reviews a previous work of hypergraph structure inference and includes a discussion of certain shortcomings their method contains. Then, Section~\ref{sec:HSLS} introduces the updated novel method HSLS which comprises of a cost function, solved via a convex optimization algorithm, and a hyperedge reduction technique. Section~\ref{sec:Experiments} experimentally evaluates the ability of HSLS to recover accurate and reasonable hypergraph structures and maintain the scalability of the method. Section~\ref{sec:Conclusion} summarizes our findings, and the Appendix contains proofs and derivations.

%% file: Sections/preliminaries.tex
\section{Preliminaries} \label{sec:Preliminaries}

A hypergraph can capture higher order relationships beyond just the pairwise. More formally, a hypergraph \(\mathcal{H} = (V(\mathcal{H}),E(\mathcal{H}))\) is made up of a set of nodes \(V(\mathcal{H})=\{v_1,v_2,\dots,v_N\}\) and set of hyperedges \(E(\mathcal{H}) = \{\mathbf{e}_1,\mathbf{e}_2,\dots,\mathbf{e}_Y\}\) where a hyperedge \(\mathbf{e}_y\) is a set that is allowed to connect any number of nodes simultaneously. The number of nodes is represented as \(N = |V(\mathcal{H})|\), and \(Y =|E(\mathcal{H})|\) is the number of hyperedges. The cardinality of a hyperedge is the number of nodes connected in the hyperedge set. The maximum cardinality \(M\) of \(\mathcal{H}\) is the largest number of nodes contained in any one hyperedge such that \(\mathbf{e}_y = \{v_i,v_j,\dots,v_l : |\mathbf{e}_y| \leq M\}\) and \(M \leq N\). Hyperedges may be assigned weights indicating the strength of relation between the connected nodes. Larger weights indicate stronger relations, and the system of defining these relationships is application dependent.

If all the hyperedges in a hypergraph have the same cardinality, then \(\mathcal{H}\) is uniform; otherwise, it is non-uniform. Figure~\ref{fig:hypergraph} contains an illustration of a non-uniform hypergraph with \(N=7\) and \(M=4\). Hypergraph \(\mathcal{H}\) can also be represented with an incidence matrix \(\mathbf{H} \in \mathbb{R}^{N \times Y}\) where each element \(H_{ij} \geq 0\). Incidence matrix \(\mathbf{H}\) can be unweighted, in which case it is a binary matrix, or weighted with scalar values between 0 and 1. If \(H_{ij}> 0\), then \(v_i \in \mathbf{e}_j\), otherwise \(v_i \not\in \mathbf{e}_j\). A hypergraph signal \(\mathbf{x} = [x_1,x_2,\dots,x_N]^\top \in \mathbb{R}^{N}\) maps each \(v_i\) to a value \(x_i\). If there are \(P\) hypergraph signal observations, a hypergraph signal matrix \(\mathbf{X} = [\mathbf{x}_1, \mathbf{x}_2, \dots, \mathbf{x}_P] \in \mathbb{R}^{N \times P}\) stores each observation column-wise. We interchangeably refer to \(\mathbf{X}\) as the hypergraph signal matrix, time-series signals, and node features.

\begin{figure}[!t]
    \centering
    \subfloat[]{\includegraphics[width=0.65\linewidth]{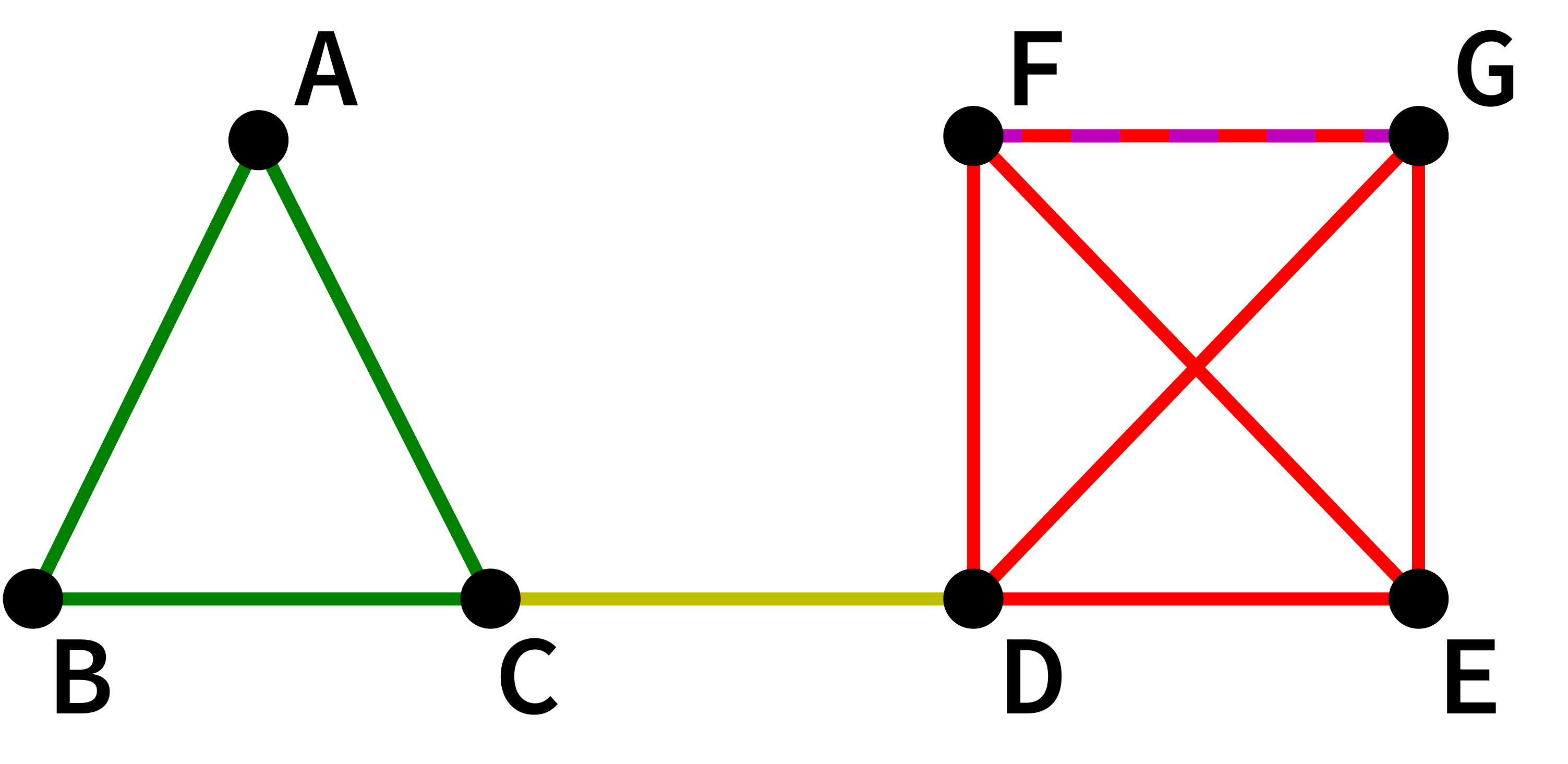}
    \label{fig:expansionsa}} \\
    \subfloat[]{\includegraphics[width=0.45\linewidth]{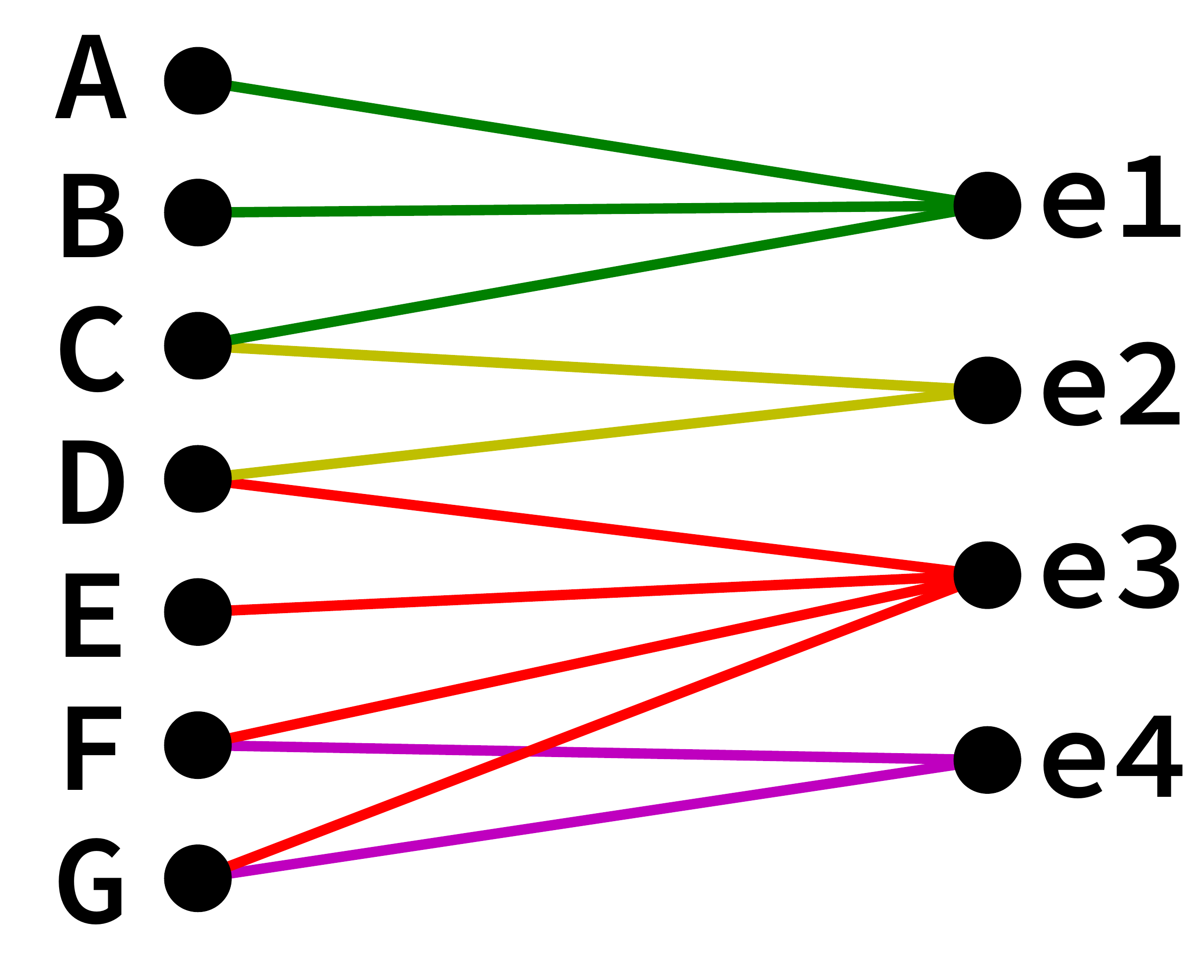}
    \label{fig:expansionsb}}
    
    \caption{Different hypergraph representations through graph expansions of the original structure in Figure~\ref{fig:hypergraph}. They are (a) clique expansion and (b) star expansion. The expansion in (a) fails to distinguish between nodes \textbf{F} and \textbf{G} in hyperedge \textbf{e3} and \textbf{e4} as the clique expansion aggregates this information into one representative edge (the red and purple line). Expansion (b) accurately captures this difference. Therefore, in the absence of \textbf{e4}, the clique expansion would remain structurally the same as in (a) while (b) would update accordingly.}
    \label{fig:expansions}
\end{figure}

A potential optimization framework for network inference~\cite{Kalofolias,MoreGraphSmoothness2} is defined as
\begin{equation} \label{eq:general}
    \underset{\mathbf{w} \in \mathcal{W}}{\text{min}} \quad \Omega(\mathbf{w}) + \theta({\mathbf{w}})
\end{equation}
where \(\mathbf{w}\) is some vector from the set of valid vectors \(\mathcal{W}\) that stores the resultant structure, \(\Omega(\mathbf{w})\) is the total variation term, and \(\theta(\mathbf{w})\) is a term that imposes further structure on the network. Kalofolias defines \(\mathbf{w}\) to contain the weights assigned to every graph edge possibility~\cite{Kalofolias}. In the context of hypergraphs, this concept can be generalized such that \(\mathbf{w} = [w_{\mathbf{e}_1}, w_{\mathbf{e}_2}, \dots,w_{\mathbf{e}_D}] \in \mathbb{R}_{\geq0}^D\) is the vector of weights assigned to every hyperedge possibility, \(D\). To clarify the use of \(D\), let there be a set \(\mathcal{K} = \{k_1,k_2,\dots,k_L\}\) defining all the cardinalities present in a hypergraph where each \(k_l \in \mathbb{Z}^+\) is a hyperedge cardinality, and \(k_l \leq M\). In the case of a hypergraph with unknown topology, but known number of nodes \(N\) and maximum cardinality \(M\), we define
\begin{equation} \label{eq:permutations}
    D = \sum_{l=1}^{L}\binom{N}{k_l}
\end{equation}
such that \(D\) is the count of every possible hyperedge a hypergraph with cardinalities \(\mathcal{K}\) could have and $L=|\mathcal{K}|$.

In graph learning, total variation \(\Omega(\mathbf{w})\) has a unified definition of \(\text{tr}(\mathbf{X}^\top\mathbf{L}\mathbf{X})\) where \(\text{tr}(\cdot)\) is the trace of a matrix, \(\mathbf{X} \in \mathbb{R}^{N \times P}\) is the graph signal matrix, and \(\mathbf{L} \in \mathbb{R}^{N \times N}\) is the graph Laplacian which stores the graph topology. Rewriting this such that the graph edge weight vector \(\mathbf{w}\) is the independent variable, the same as used by Kalofolias, graph total variation can be written as
\begin{equation}\label{eq:graph_total_variation}
    \Omega(\mathbf{w}) = \mathbf{z}^\top \mathbf{w}
\end{equation}
where each element of the pairwise distance vector, \(\mathbf{z}\), is \(z_d = ||\mathbf{X}_{i,:} - \mathbf{X}_{j,:}||_2^2\) and \(\mathbf{X}_{i,:}^\top,\mathbf{X}_{j,:}^\top \in \mathbb{R}^P\) are graph time-series signals on nodes \(v_i\) and \(v_j\), respectively, corresponding to row \(i\) and \(j\) of \(\mathbf{X}\). In parallel, hypergraph total variation can be represented in the same way. In this case, \(\mathbf{w} \in \mathbb{R}^D\) holds the hyperedge weights and \(\mathbf{z} = [z_{\mathbf{e}_1}, z_{\mathbf{e}_2}, \dots, z_{\mathbf{e}_D}] \in \mathbb{R}^D\) holds some form of distance measure for each potential hyperedge. Since hyperedges are allowed to connect an arbitrary number of nodes together, the realization of the pairwise distance vector needs to be updated to a generalized distance measure. This notion has inspired several kinds of hyperedge distance definitions. With each new calculation stored in \(\mathbf{z}\), there is a new hypergraph total variation measure \(\Omega(\mathbf{w})\).

Nguyen and Mamitsuka~\cite{SmoothnessTable} show that the majority of total variations have a unified form of
\begin{equation} \label{eq:unified_form}
    \Phi(\mathbf{X}) = T_{\mathbf{e}_d \in E(\mathcal{H})}(w_{\mathbf{e}_d} \cdot t_{v_i,v_j \in \mathbf{e}_d} (s(\mathbf{X}_{i,:},\mathbf{X}_{j,:})))
\end{equation}
where \(s(\mathbf{X}_{i,:},\mathbf{X}_{j,:})\) is a kind of measure of difference between node time-series signals. The function, \(t(\cdot)\), then combines these measures across all pairings of nodes \(v_i,v_j \in \mathbf{e}_d\) into the distance measure which is stored in \(\mathbf{z}\). This can then be scaled by the hyperedge weight \(w_{\mathbf{e}_d}\). Finally, \(T(\cdot)\) combines all measures across all hyperedges from \(E(\mathcal{H})\). In its entirety, the unified smoothness term is defined as \(\Phi(\mathbf{X})\). We find that the smoothness measurements we use follow the unified form of \eqref{eq:unified_form} and therefore are valid hypergraph total variations.

It is worth recognizing that there is a wide selection of total variation definitions on hypergraphs. The list is rather expansive, especially considering generalized frameworks like that of \eqref{eq:unified_form} and the between-hyperedge  $\ell_p$ norm~\cite{SmoothnessTable} defined as
\begin{equation} \label{eq:lp_tv}
    \left(\sum_{\mathbf{e}_d \in E(\mathcal{H})}\sum_{v_i,v_j \in \mathbf{e}_d} ||\mathbf{X}_{i,:} - \mathbf{X}_{j,:}||_p \right)^{\frac{1}{p}},
\end{equation}
where \eqref{eq:lp_tv} is similar to the \textit{p}-regularizer used in undirected graph works~\cite{GraphLP1,GraphLP2}. There are many combinations of functions that can be substituted into \eqref{eq:unified_form} and many values of $p$ that can be used in \eqref{eq:lp_tv}. To this end, we have chosen a group of valid total variation terms that are able to be cast in the form of \eqref{eq:graph_total_variation}. The chosen smoothness terms are by no means an exhaustive list; however, they do cover a diverse range of linear and nonlinear forms and common $\ell_p$ norms. Furthermore, all definitions fit \eqref{eq:unified_form}, and some also satisfy \eqref{eq:lp_tv}.

We now introduce the four total variation measures used in this work, all of which are convex operations. A first hypergraph total variation of interest to learning hypergraphs is the {\em Sum-Square} variation found in the work from Agarwal et al.~\cite{StarCliqueSimilarity} defined according to
\begin{equation} \label{eq:smooth1}
    \sum_{\mathbf{e}_d \in E(\mathcal{H})} \sum_{v_i,v_j \in \mathbf{e}_d} w_{\mathbf{e}_d} \cdot ||\mathbf{X}_{i,:} - \mathbf{X}_{j,:}||_2^2,
\end{equation}
which uses the squared Euclidean distances between node time-series. The summations are across all pairs \(v_i\) and \(v_j\) in the hyperedge of interest, \(\mathbf{e}_d\), and then across all hyperedges, \(\mathbf{e}_d \in E(\mathcal{H})\). When \(M=2\), this smoothness measure decomposes to the traditional means of measuring total variation in graphs, and can be derived by either star or clique expansions.  Figure~\ref{fig:expansionsa} and~\ref{fig:expansionsb} contain an example clique and star expansion, respectively, of the hypergraph of Fig.~\ref{fig:hypergraph}. The clique expansion is considered only as an approximation as it cannot represent non-isomorphic hypergraphs with the same clique expansion~\cite{isomorphic}. More concretely, if there are two hyperedges where one has at least two nodes that are the subset of the other, then the clique expansion would have at least one edge with a multiplicity of 2. It is standard for the clique expansion to aggregate edges with multiplicity greater than 1 into a single representative edge~\cite{isomorphic,Clique1,Clique2}, in which case the distinction between hyperedges is lost. Alternatively, the star expansion demonstrates the ability to accurately represent all hyperedges, and as such is considered a better representation. This concept is demonstrated in the example of Fig.~\ref{fig:expansions}.

The second hypergraph total variation of interest from Nguyen and Mamitsuka~\cite{SmoothnessTable}, noted to be associated with the clique expansion and align with the framework of \eqref{eq:lp_tv} when $p=1$, is the {\em Sum-Absolute} variation defined as
\begin{equation} \label{eq:smooth2}
     \sum_{\mathbf{e}_d \in E(\mathcal{H})}\sum_{v_i,v_j \in \mathbf{e}_d} w_{\mathbf{e}_d} \cdot ||\mathbf{X}_{i,:} - \mathbf{X}_{j,:}||_1,
 \end{equation}
which is similar to \eqref{eq:smooth1} except the difference measure is now the absolute value. Both \eqref{eq:smooth1} and \eqref{eq:smooth2} sum the differences between node time-series for all \(v_i,v_j \in \mathbf{e}_d\). This property is not advantageous when combined with a minimization problem such as \eqref{eq:general}. The main issue is that as the cardinality of a hyperedge increases, there are more pairs of \(v_i,v_j\) and, therefore, more summations of time-series differences increasing the total variation. With no sort of normalization or averaging, this causes a magnitude imbalance between the contribution of hyperedges with mixed cardinalities to the overall total variation.

With the issues of \eqref{eq:smooth1} and \eqref{eq:smooth2} in mind, another hypergraph total variation is found in the work of Hein et al. \cite{OneOfTheSmoothnessTerms} who propose the {\em Max-Absolute} variation defined by
\begin{equation} \label{eq:smooth3}
    \sum_{\mathbf{e}_d \in E(\mathcal{H})} w_{\mathbf{e}_d} \cdot \underset{v_i,v_j \in \mathbf{e}_d}{\text{max}} ||\mathbf{X}_{i,:} - \mathbf{X}_{j,:}||_1,
\end{equation}
which injects some nonlinearity by taking the maximum absolute difference between node time-series within a hyperedge. This total variation also follows the framework of \eqref{eq:lp_tv} where the inner summation is allowed to be replaced with the nonlinear maximum operation, and $p=1$. The use of the max difference avoids the magnitude imbalance issue and provides an interesting notion of hypergraph smoothness. This implies that the total variation contribution of a hyperedge is governed by the two nodes with the least similarity, or largest variance.

The use of the absolute difference in a minimization problem may be problematic as any perturbation to the smoothness calculation could result in changes that are too small to have a meaningful effect. To counteract this, we introduce the last hypergraph total variation to be the {\em Max-Square} variation defined by
\begin{equation} \label{eq:TangSmoothness}
    \sum_{\mathbf{e}_d \in E(\mathcal{H})} w_{\mathbf{e}_d} \cdot \underset{v_i,v_j \in \mathbf{e}_d}{\text{max}} ||\mathbf{X}_{i,:} - \mathbf{X}_{j,:}||_2^2,
\end{equation}
which was implemented by Tang et al.~\cite{Tang} and derived based on a probabilistic model with the star expansion. The introduction of the squared difference with the max term enhances the changes to the hypergraph structure more so than the absolute difference of the Max-Absolute variation.

%% file: Sections/HGSI.tex
\section{Hypergraph Structure Inference Under Smoothness Prior (HGSI)} \label{sec:HGSI}
Tang et al.~\cite{Tang} proposed the hypergraph structure inference method HGSI to make use of the Max-Square variation, through \(\Omega(\mathbf{w}) = \mathbf{z}^\top \mathbf{w}\), and includes the minimization
\begin{equation}\label{eq:TangOptimization}
    \underset{\mathbf{w}}{\text{min}} \ \mathbf{z}^\top \mathbf{w} - \mathbf{1}^\top \text{log}(\mathbf{w}) + ||\mathbf{w}||_1,
\end{equation}
recovering hyperedge weights, \(\mathbf{w}\), based on a probabilistic model describing hypergraph structure and node feature relations.

The objective function for HGSI is convex with the optimal closed form solution
\begin{equation}\label{eq:ClosedForm}
    \mathbf{w}^\star = \frac{1}{\mathbf{z} + 1},
\end{equation}
calculated element-wise on \(\mathbf{z}\) where \(\mathbf{w}^\star \in (0,1]^{\bar{D}}\). The recovery of \(\mathbf{w}^\star\) is the equivalent of learning the hypergraph structure since it contains the weights for each potential hyperedge, indicating a level of confidence that each hyperedge is in the hypergraph.

We note that the length of \(\mathbf{w}^\star\) is not \(D\), but rather \(\bar{D}\). It is recognized that computing \(\mathbf{w}^\star\) is time inefficient and, as \(N\) and \(M\) increase, often completely infeasible due to the vast number of hyperedge combinations. So, to limit the number of hyperedge possibilities, HGSI employs the use of a K-NN based method for possibility reduction. Specifically, given the set of cardinalities \(\mathcal{K}\), determine the \(k_l-1\) nearest neighbors of each node based on time-series observations. Then, let the combinations of the original nodes and their nearest neighbors be the new set of possible hyperedges where \(\bar{D} = NL\) is the upper bound of hyperedge possibilities in the set reduced by K-NN for $L = |\mathcal{K}|$. This is considered an upper bound since the use of K-NN could produce repeat hyperedges which would eventually be aggregated into one copy.

HGSI has advantages when it comes to scalability through the use of a closed form solution and the hyperedge reduction technique. However, both of these factors also cause certain shortcomings as they relate to assumptions on hypergraph structure. The first is that the format of \eqref{eq:ClosedForm} improperly constrains the set of learned hyperedges in a non-uniform hypergraph structure as a factor of \(\mathbf{z}\). If there is some non-uniform hypergraph with hyperedges \(\mathbf{e}_i\) and \(\mathbf{e}_j\), then we can say if \(|\mathbf{e}_i| < |\mathbf{e}_j|\) and \(\mathbf{e}_i \subset \mathbf{e}_j\), then \(w_{\mathbf{e}_j} \leq w_{\mathbf{e}_i}\). This means that higher cardinality hyperedges will be assigned a weight that is either the same or less than the weight of some lower cardinality hyperedge subset. Therefore, if the selection of hyperedges prioritizes choosing those with the largest weights first, then the closed form solution implies that lower cardinality hyperedges could be assigned greater importance. This is an assumption that may not hold in all cases, and establishes more priors beyond that of smoothness.

The second shortcoming of HGSI is in the method of hyperedge possibility reduction. The use of K-NN provides a notion of local smoothness on the hypergraph by focusing on one node at a time and its nearest neighbors with no other outside information. While local smoothness is certainly a related concept, the use of K-NN fails to capture any global smoothness which is of greater interest when establishing a smoothness prior. This reduction method exaggerates the overlap between the set of true hyperedges and those produced by K-NN. HGSI ends up discarding valid hyperedge possibilities under the local smoothness assumption, as we will demonstrate in Sec.~\ref{sec:Experiments}.

%% file: Sections/HSLS.tex
\section{Hypergraph Structure Learning with Smoothness (HSLS)} \label{sec:HSLS}

We now formulate our own method to recover the hypergraph structure to overcome the shortcomings of previous works. Our method, which we coin as Hypergraph Structure Learning with Smoothness (HSLS), centers around learning the weight vector \(\mathbf{w} \in \mathbb{R}^D\) for an initial \(D\) hyperedge possibilities. It is important to emphasize that the optimization procedure that will be implemented is from an existing source that is fit to our model. Although the problem will be shown as being solved using a standard convex optimization algorithm, the cost function is newly implemented. This cost function encodes domain-specific structure that has not been captured in prior formulations, and it is one source of the methodological novelty. Furthermore, while the employed convex optimization algorithm is pre-existing, since the cost function is a new interpretation, this constitutes a unique optimization setup.

Using the work of Tang et al. in \eqref{eq:TangOptimization} as reference, we define our total variation as \(\Omega(\mathbf{w}) = \mathbf{z}^\top \mathbf{w}\). In the same experimental context, the elements of \(\mathbf{z}\) will vary depending on the smoothness term selected, but it always contains the value of the time-series difference measures. So, for the Sum-Square total variation, \(\mathbf{z}\) holds the sum of the squared differences, while for Max-Absolute, it holds the maximum of the absolute differences, and so on for the other terms. The format of \(\Omega(\mathbf{w})\) provides the ability to substitute in any of the selected smoothness terms creating a robust optimization. Consequently, we can evaluate in the same setting which terms perform better. Since all total variation terms can fit this convex format, the optimization we pose will also maintain convexity. Furthermore, we reiterate that we invoke a smoothness prior for the signals on the hypergraph structure which \(\Omega(\mathbf{w})\) enforces.

Revisiting the general form of \eqref{eq:general}, the use of \(\theta(\mathbf{w})\) includes other terms that impose further structure on the hypergraph. We will be using a degree positivity and sparsity term in our \(\theta(\mathbf{w})\). After substituting the terms into \eqref{eq:general}, the final formulation of our convex cost function is
\begin{equation} \label{eq:optimization}
    \underset{\mathbf{w} \in \mathcal{W}}{\text{min}} \quad \mathbf{z}^\top \mathbf{w} - \alpha \mathbf{1}^\top \log(\mathbf{Sw}) + \beta ||\mathbf{w}||^2_2
\end{equation}
where \(\log(\cdot)\) is the natural logarithm taken element-wise and \(||\cdot||_2^2\) is the squared $\ell_2$ norm of a vector. The \(\alpha\) and \(\beta\) are positive scalar parameters used to control the influence of their corresponding function. The matrix \(\mathbf{S} \in \mathbb{R}^{N \times D}\) performs a linear transformation on \(\mathbf{w}\) to convert \(\mathbf{Sw} = \mathbf{d}\). Here, \(\mathbf{d} \in \mathbb{R}^N\) is the degree vector containing the degree of each node which is the sum off all the weights corresponding to hyperedges that contain \(v_i\). Hence, the negative log term acts on degree positivity by allowing nodes to still have a degree close to zero, but remain positive. Also, since it is acting on \(\mathbf{d}\), the \(\alpha\) parameter can be used to control the magnitude of the node degrees. The squared $\ell_2$ norm is the sparsity term which controls the spread of the weights across the hyperedges. Similar to \(\alpha\), the \(\beta\) parameter is used to control the level of influence of sparsity on the hypergraph structure.

Here, we note the key differences between \eqref{eq:TangOptimization} and \eqref{eq:optimization}. First, as previously mentioned, our smoothness term is no longer fixed, but allowed to be substituted with the selected total variation definitions. Second, the degree positivity term is now based on \(\mathbf{d}\) instead of strictly \(\mathbf{w}\). The inclusion of \(\mathbf{d}\) is used in the tensor based hypergraph learning work of Pena-Pena et al.~\cite{Karelia} and the graph learning work of Kalofolias~\cite{Kalofolias}. The use of the degree vector is critical to avoid the closed form solution of \eqref{eq:ClosedForm}, which in turn overcomes the issue of hyperedges being assigned unbalanced importance. Third, we change the sparsity term to be the squared $\ell_2$ norm which makes the optimization more sensitive to smaller perturbations in \(\mathbf{w}\). Finally, we include the parameters \(\alpha\) and \(\beta\) for greater control over the influence of each structural parameter.

Since our method uses \(\mathbf{d}\), there is no longer a straightforward closed form solution like that of \eqref{eq:ClosedForm}. Instead, \eqref{eq:optimization} can be solved using a convex primal-dual algorithm which avoids the issue of hyperedges with varying cardinalities being assigned weighted importance. Here, it is already cast in the appropriate convex form according to Komodakis and Pesquet~\cite{Duality} which is
\begin{equation} \label{eq:convex}
    \underset{\mathbf{w} \in \mathcal{W}}{\text{min}} \quad f(\mathbf{w}) + g(\mathbf{Sw}) + h(\mathbf{w})
\end{equation}
where \(h(\mathbf{w})\) must be differentiable with a positive Lipschitzian constant \(\zeta\). The terms of \eqref{eq:optimization} parallel those of \eqref{eq:convex} as follows:
\begin{align}
    f(\mathbf{w}) &= \mathbf{z}^\top \mathbf{w} + \mathbbm{1}\{\mathbf{w} \geq 0\}, \\
    g(\mathbf{Sw}) &= -\alpha \mathbf{1}^\top \log(\mathbf{Sw}), \\
    h(\mathbf{w}) &= \beta ||\mathbf{w}||^2_2.
\end{align}
The only new term we have introduced is the indicator function \(\mathbbm{1}\{\cdot\}\) defined as
\begin{align}
\mathbbm{1}\{w_{\mathbf{e}_d} \geq 0\} = \left\{
\begin{aligned}
    &1 \quad & ,w_{\mathbf{e}_d} \geq 0\\
    &+\infty \quad &  ,w_{\mathbf{e}_d} < 0
\end{aligned}
\right.
\end{align}
operating element-wise on \(\mathbf{w}\). The inclusion of the indicator function is what enforces \(\mathbf{w} \in \mathcal{W}\) as weights can only be non-negative real values.

From Komodakis and Pesquet~\cite{Duality}, we have chosen to use the convex Forward-Backward-Forward (FBF) based primal-dual algorithm to perform the minimization. The FBF algorithm requires the use of the gradient of \(h(\cdot)\), the proximal operator of \(f(\cdot)\), and the proximal operator of the conjugate of \(g(\cdot)\) denoted as \(g^*(\cdot)\). These operators are defined as
\begin{align}
    \label{eq:gradh}
    \nabla h(\mathbf{x}) &= 2 \beta \mathbf{x}, \\
    \label{eq:proxf}
    (\text{prox}_{\gamma f}(\mathbf{x}))_i &= \text{max}(0, x_i - \gamma z_{{\mathbf{e}}_i}), \\
    \label{eq:proxg}
    (\text{prox}_{\gamma g^*}(\mathbf{x}))_i &= x_i - \gamma\left(\frac{y_i + \sqrt{y_i^2 + 4 \alpha/\gamma}}{2}\right),
\end{align}
where the derivations can be found in Appendix~\ref{sec:AppendixA} and~\ref{sec:AppendixB}.
The proximal operators are written in element-wise form to clarify how the max, square, and square root operations are being performed, and each \(z_{{\mathbf{e}}_i}\) is an element from the distance vector \(\mathbf{z}\). The \(\gamma\) is a constant set during the optimization process. For the proximal operator on \(g^*\), we substituted \(y_i = x_i / \gamma\).

Now, in reference to the size of the hyperedge search space, the use of \(D\) hyperedge possibilities, which is every potential hyperedge combination, is incredibly costly as \(N\) and \(|\mathcal{K}|\) increase. HGSI attempts to counteract this by reducing the hyperedge selection set using each node's \(k_l - 1\) nearest neighbors. This approach is not very effective as many potential valid hyperedges are removed from consideration. We demonstrate these concepts in Sec.~\ref{sec:Experiments} where we find that, after a certain number of nearest neighbors, all hyperedges of a ground truth hypergraph can be included in the valid hyperedge set.
\begin{algorithm}[!t]
\caption{Method of Hyperedge Reduction for HSLS.}
\begin{algorithmic}[1]
    \STATE \textbf{Input:} \(\mathbf{X}, {\mathcal{K}}, {\mathcal{R}}, V(\mathcal{H}), \)
    \STATE \textbf{Output:} \(\hat{E}(\mathcal{H}), \hat{D}=|\hat{E}(\mathcal{H})|\)
    \FOR{\(l = 1:L\)}
        \FOR{\(i = 1:N\)}
            \STATE \(\mathcal{J}_{v_i} = KNN(\mathbf{X}, v_i\in V(\mathcal{H}), r_l \in  {\mathcal{R}})\) set of $r_l$ nodes that are nearest neighbors of $v_i$ in $\mathbf{X}$, $v_i$ excluded.
            \STATE \(\mathcal{C}_{v_i} = \{\{v_i,v_{j_1},v_{j_2},\dots,v_{j_{k_l - 1}}\} : v_{j_p}, v_{j_q} \in \mathcal{J}_{v_i}, v_{j_p} \neq v_{j_q}, k_l \in \mathcal{K}\}\) all unique combinations of \(v_i\) with \(k_l-1\) nodes from \(\mathcal{J}_{v_i}\).
        \ENDFOR
        \STATE $\hat{E}_{k_l}(\mathcal{H}) = \bigcup_{i=1}^{N} \mathcal{C}_{v_i}$ reduced set of hyperedges for cardinality $k_l$.
    \ENDFOR
    \STATE  $\hat{E}(\mathcal{H}) = \bigcup_{l=1}^{L} \hat{E}_{k_l}(\mathcal{H})$ reduced set of hyperedges for all cardinalities.
\end{algorithmic}
\label{alg:hyperedge_reduction}
\end{algorithm}

To this end, we adopt an approach where a variable number of nearest neighbors can be selected per cardinality in $\mathcal{K}$. This method of hyperedge reduction can be found in Algorithm~\ref{alg:hyperedge_reduction} and is detailed here. We introduce $\mathcal{R} = \{r_1,r_2,\dots,r_L\}$ such that $r_l \geq k_l - 1$ is a lower bound, and $|\mathcal{R}| = |\mathcal{K}|$. The elements of set $\mathcal{R}$ are considered the number of nearest neighbors to be taken per node with reference to $\mathcal{K}$. Clarifying this point, $\mathcal{K}$ is the set of cardinalities in a hypergraph. Each $r_l$ corresponds to one of these cardinalities $k_l$ specifying the number of nearest neighbors in $\mathbf{X}$ to be taken. The $r_l$ nearest neighbors of node $v_i$ get stored in $\mathcal{J}_{v_i}$. We define function $KNN(\cdot)$ to produce the $r_l$ nodes that are the nearest neighbors of $v_i$ based on Euclidean distances of the time-series in $\mathbf{X}$. We then take every unique combination of $v_i$ with $k_l-1$ nodes from $\mathcal{J}_{v_i}$ and store as set $\mathcal{C}_{v_i}$. The union of all $\mathcal{C}_{v_i}$ produces $\hat{E}_{k_l} (\mathcal{H})$ the set of hyperedges for one cardinality, and the union of $\hat{E}_{k_l} (\mathcal{H})$ across all cardinalities is $\hat{E}(\mathcal{H})$, the final reduced hyperedge set. This hyperedge set has an upper bound size
\begin{equation}
    \hat{D} = \sum_{l=1}^{L} \binom{r_l}{k_l-1},
\end{equation}
which is considered an upper bound since the use of K-NN could produce repeat hyperedges which are aggregated into one copy, similar to the discussion of HGSI.

To make this more concrete, consider the following example. Let $\mathcal{K} = \{3,4\}$ and $\mathcal{R} = \{15,10\}$ for $N=10$. Then, Algorithm~\ref{alg:hyperedge_reduction} would start with $r_1 = 15$ which corresponds to cardinality $k_1 = 3$. Then, considering only $v_1$ currently, its $r_1=15$ nearest neighbors from $\mathbf{X}$ (excluding itself) are found and labeled set $\mathcal{J}_{v_1}$. From the $r_1=15$ nodes in $\mathcal{J}_{v_1}$, $v_1$ is paired with every unique combination of $k_1 - 1 = 2$ of these nearest neighbor nodes. This produces $\binom{r_1}{k_1}=105$ combinations of order $k_1=3$ hyperedges between $v_1$ and its set of $r_1=15$ nearest neighbors from $\mathcal{J}_{v_1}$, and this is stored in $\mathcal{C}_{v_1}$. This process is repeated for every other node to get $\hat{E}_{k_1}(\mathcal{H}) = \bigcup_{i=1}^{10} \mathcal{C}_{v_i}$, and then again for every other cardinality in $\mathcal{K}$ for final output $\hat{E}(\mathcal{H}) = \bigcup_{l=1}^{2} \hat{E}_{k_l}(\mathcal{H})$.

\begin{algorithm}[!t]
\caption{FBF Primal-Dual Algorithm for solving \newline Equation~\eqref{eq:optimization}.}
\begin{algorithmic}[1]
    \STATE \textbf{Input:} \(\alpha, \beta, \eta, \gamma 
\text{ sequence}, i_{max}, \mathbf{z}, \mathbf{S} \in \mathbb{R}^{N \times \hat{D}}, \mathbf{w}^0 \in \mathbb{R}_{\geq 0}^{\hat{D}}, \mathbf{d}^0 \in \mathbb{R}_{\geq 0}^N\)
    \STATE \textbf{Output:} \(\mathbf{w}^{i_{end}},\mathbf{d}^{i_{end}}\)
    \FOR{\(i = 0, 1, ..., i_{max}\)}
        \STATE Set \(\gamma^i\)
        \STATE \(\mathbf{y}^{i} = \mathbf{w}^i - \gamma^i(\nabla h(\mathbf{w}^i) + \mathbf{S}^\top \mathbf{d}^i)\)
        \STATE \(\hat{\mathbf{y}}^{i} = \mathbf{d}^i + \gamma^i \mathbf{Sw}^i\)
        \STATE \(\mathbf{p}^{i} = \text{prox}_{\gamma^i f}(\mathbf{y}^{i})\)
        \STATE \(\hat{\mathbf{p}}^{i} = \text{prox}_{\gamma^i g^*} (\hat{\mathbf{y}}^{i})\)
        \STATE \(\mathbf{q}^{i} = \mathbf{p}^{i} - \gamma^i(\nabla h(\mathbf{p}^{i}) + \mathbf{S}^\top \hat{\mathbf{p}}^{i})\)
        \STATE \(\hat{\mathbf{q}}^{i} = \hat{\mathbf{p}}^{i} + \gamma^i \mathbf{S} \mathbf{p}^{i}\)
        \STATE \(\mathbf{w}^{i+1} = \mathbf{w}^i - \mathbf{y}^{i} + \mathbf{q}^i\)
        \STATE \(\mathbf{d}^{i+1} = \mathbf{d}^i - \hat{\mathbf{y}}^{i} + \hat{\mathbf{q}}^i\)
        \IF{$||\mathbf{w}^{i+1} - \mathbf{w}^i||_2^2 / ||\mathbf{w}^i||_2^2 < \eta$ \textbf{and}\\
            \hspace{0.6em} $||\mathbf{d}^{i+1} - \mathbf{d}^i||_2^2 / ||\mathbf{d}^i||_2^2 < \eta$}
            \STATE \textbf{break}
        \ENDIF
    \ENDFOR
\end{algorithmic}
\label{alg:HSLS}
\end{algorithm}

We can now define the minimization solver for \eqref{eq:optimization} as shown in Algorithm~\ref{alg:HSLS}. Most of the inputs are the same as those included in \eqref{eq:optimization} where \(\mathbf{w}\) is the primal variable and \(\mathbf{d}\) is the dual variable. With the inclusion of Algorithm~\ref{alg:hyperedge_reduction} as a pre-processing step for Algorithm~\ref{alg:HSLS}, we updated the appropriate dimensions with \(\hat{D}\). The superscripts indicate which iteration of the vectors are being used as they are updated on each pass. The parameter, \(\gamma_i\), is effectively the learning rate, but in every forward pass, it is updated according to a sequence determined by the Lipschitz constant, \(\zeta\). See Appendix~\ref{sec:AppendixC} for details on the derivation of this sequence. The \(\eta\) is the threshold for the stopping condition if the max iteration \(i_{max}\) is not reached first. The convergence rate is $\mathcal{O}(1/(i+1))$ for iteration $i$, and the overall complexity is $\mathcal{O}(N\hat{D})$. The proximal operator of function $f$ has complexity $\mathcal{O}(\hat{D})$ and the proximal operator of function $g^*$ has complexity $\mathcal{O}(N)$. Following the guidelines of Komodakis and Pesquet~\cite{Duality}, HSLS is guaranteed to converge to the approximate global minimum.

%% file: Sections/experiments.tex
\section{Experiments}\label{sec:Experiments}
We now evaluate the performance of our model in experimental settings\footnote{All source code used for our experiments is publicly available at https://github.com/Ben-Brown-Code/Scalable-Hypergraph-Structure-Learning-with-Diverse-Smoothness-Priors}. First, we test on co-authorship networks with both uniform and non-uniform structures. In this scenario, the underlying hypergraph structure is known and will serve as our ground truth. The purpose of this experiment is to evaluate the accuracy with which our optimization combined with various total variations can recover a hypergraph topology. The results are a report of F1-Score, Precision, and Recall between the learned hypergraph and the ground truth. Furthermore, this experiment serves to demonstrate the shortcomings of HGSI using K-NN as a method of hyperedge possibility reduction, as mentioned in Sec.~\ref{sec:HGSI}, and how that method can be improved upon. We then test on a set of lung cancer mortality data across the U.S. This experiment is designed to evaluate the effectiveness of each total variation in producing a smooth hypergraph structure. This data has no ground truth, but a comparison to a baseline hypergraph is used to highlight the emphasis our algorithm puts on selecting hyperedges that create a globally smooth topology. Finally, we apply HSLS to a hyperspectral image classification task to improve the results of a baseline method and further the total variation analysis.

\subsection{Co-authorship Networks}
We experiment on two real world co-authorship networks: DBLP and Cora \cite{CoraAndDBLP}. For each dataset, nodes are considered authors, and a hyperedge groups nodes if the nodes' corresponding authors have written a paper together. We use a subset of the overall datasets to experiment on, and these subsets are the same that Pena-Pena et al.~\cite{Karelia} used. Therefore, we can compare our learning method and smoothness terms to their tensor based method PDL-HGSP. We also compare results with the HGSI method from Tang et al.~\cite{Tang} and the K-NN approach in the node feature space from Gao et al.~\cite{HGNN+}. The experiment is performed on both a uniform and non-uniform cardinality hypergraph from both DBLP and Cora. In the uniform scenarios, the set of cardinalities \(\mathcal{K} = \{3\}\). In the non-uniform scenarios, \(\mathcal{K} = \{2,3\}\). Table \ref{tab:NetworkNodesEdges} details the structural parameters of the ground truth hypergraphs used.

\begin{table}[!t]
    \captionsetup{justification=centering}
    \caption{Number of nodes and hyperedges for experimental co-authorship networks.}
    \label{tab:NetworkNodesEdges}
    \centering
    \begin{tabular}{rcc}
                              & \(|V(\mathcal{H})|\)  & \(|E(\mathcal{H})|\) \\
        \midrule
        \textbf{DBLP (Uniform)}        & 40                     & 22                                 \\
        \textbf{DBLP (Non-Uniform)}    & 58                     & 39                                 \\
        \textbf{Cora (Uniform)}        & 21                     & 10                                 \\
        \textbf{Cora (Non-Uniform)}    & 61                     & 48
    \end{tabular}
\end{table}

\begin{figure*}[!t]
    \centering
    \includegraphics[width=0.75\linewidth]{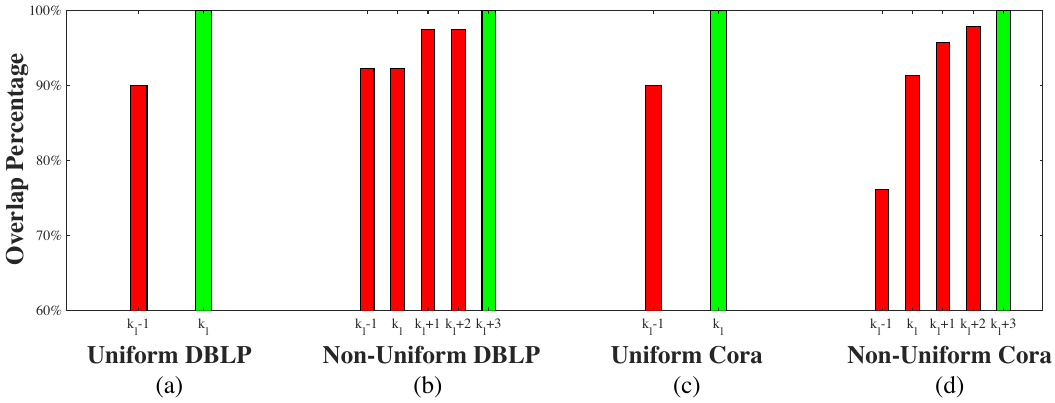}
    
    \caption{Bar graphs of overlap percentage between \(E_{truth}(\mathcal{H})\) and \(E_{KNN}(\mathcal{H})\). Red bars indicate the percentage is less than \(100\%\). Green bars indicate the overlap percentage has reached \(100\%\). The x-axis is labeled in terms of the number of nearest neighbors used dependent on \(k_l\), the cardinalities from set \(\mathcal{K}\). Each bar plot corresponds to one of the co-authorship networks: (a) Uniform DBLP, (b) Non-Uniform DBLP, (c) Uniform Cora, and  (d) Non-Uniform Cora.}
    \label{fig:OverlapPercentage}
    
\end{figure*}

Now, with regard to signals on the DBLP and Cora datasets, there is no given set of time-series observations. Since our method is based upon a smoothness prior, we first need to generate smooth signals across the network structures. We adopt the same process introduced by Tang et al.~\cite{Tang} who use a multivariate Gaussian distribution for smooth hypergraph signals, similar to processes used in graph signal processing~\cite{Dong}. Using the Laplacian matrix of the hypergraph star expansion \(\mathbf{L}_s\), the signal distribution can be modeled as
\begin{equation} \label{eq:distribution}
    [\mathbf{x}_\mathbf{V}^\top, \mathbf{x}_\mathcal{E}^\top]^\top \sim \mathcal{N}(\mathbf{0}, \mathbf{L}_s^\dagger)
\end{equation}
where \(\mathbf{L}_s^\dagger\) is the pseudo-inverse of \(\mathbf{L}_s\), \(\mathbf{0}\) is a vector of all zeros, \(\mathbf{x}_\mathbf{V} \in \mathbb{R}^N\) is the signal across hypergraph nodes, and \(\mathbf{x}_\mathcal{E} \in \mathbb{R}^{Y}\) is the signal across the hyperedges. After sampling from the multivariate Gaussian distribution, we only kept \(\mathbf{x}_\mathbf{V}\) as one hypergraph signal observation, making this generative process resemble marginalization over a Gaussian Markov Random Field induced through sampling. We stored all hypergraph signal observations in \(\mathbf{X}\). We also add some noise along the diagonal of \(\mathbf{L}_s\) on the order of \(1 \times 10^{-3}\). We note that, in this context, \(\mathbf{L}_s\) is the Laplacian of the star expansion of the ground truth hypergraphs from both the DBLP and Cora datasets and is a known quantity. We used \(P = 250\) hypergraph signal observations to produce \(\mathbf{X}\) which was kept constant across all learning methods per scenario.

We started with an analysis of the overlap between the K-NN selected hyperedges and the ground truth for all co-authorship networks. Using \(\mathcal{K}\), we selected the \(k_l-1\) nearest neighbors of each node from \(\mathbf{X}\) to form hyperedges. We then computed the overlap percentage between the set of ground truth hyperedges and the K-NN hyperedges. By this we mean that, for each hyperedge in the ground truth set \(E_{truth}(\mathcal{H})\), we determined if it was present in the K-NN hyperedge set \(E_{KNN}(\mathcal{H})\). If \(E_{truth}(\mathcal{H}) \subseteq E_{KNN}(\mathcal{H})\)  then this is a \(100\%\) overlap rate. We then increased the number of nearest neighbors by one and recomputed the overlap percentage for the new, larger K-NN set. This was repeated for a steadily increasing count of nearest neighbors. The computation of \(E_{KNN}(\mathcal{H})\) was performed using Algorithm~\ref{alg:hyperedge_reduction}, where $\mathcal{K} = \{3\}$ and $\mathcal{K} = \{3,4\}$ in the uniform and non-uniform scenarios, respectively, and $\mathcal{R}$ was continuously updated with the increasing nearest neighbor count starting at $k_l-1$. The results are shown in the bar graphs of Fig.~\ref{fig:OverlapPercentage}.

From the results of Fig.~\ref{fig:OverlapPercentage}, we note that in the uniform hypergraph scenarios, the overlap percentage starts at a higher threshold than that of the non-uniform cases. This can be attributed to two main factors. The first is that the uniform scenarios have less \(|E(\mathcal{H})|\) than the non-uniform scenarios. The second is that a mixed cardinality hypergraph has a much larger set of possible hyperedges making it less straightforward to select valid hyperedges. In the uniform cases, it takes significantly less neighbors in the feature space to achieve  \(E_{truth}(\mathcal{H}) \subseteq E_{KNN}(\mathcal{H})\) than that of the non-uniform cases. In fact, the minimum number of neighbors for this in the uniform and non-uniform scenarios is \(k_l\) and \(k_{l}+3\), respectively. The use of \(k_l-1\) nearest neighbors always falls below 100\% overlap, demonstrating the limitation of the method of HGSI since, after the reduction of hyperedge possibilities, it is impossible for it to reproduce a hypergraph that completely corresponds to the ground truth. We also find that there is indeed a reasonable point where 100\% overlap can be achieved, therefore validating our choice to incorporate a K-NN hyperedge reduction method with number of neighbors greater than \(k_l-1\). In the rest of the co-authorship network experiments, we used $\mathcal{R} = \{3\}$ in the uniform and $\mathcal{R} = \{5,6\}$ in the non-uniform pre-processing with Algorithm~\ref{alg:hyperedge_reduction}. This corresponds to the number of neighbors that produced 100\% overlap in Fig.~\ref{fig:OverlapPercentage}.

\begin{figure}[!t]
    \centering
    \includegraphics[width=\linewidth]{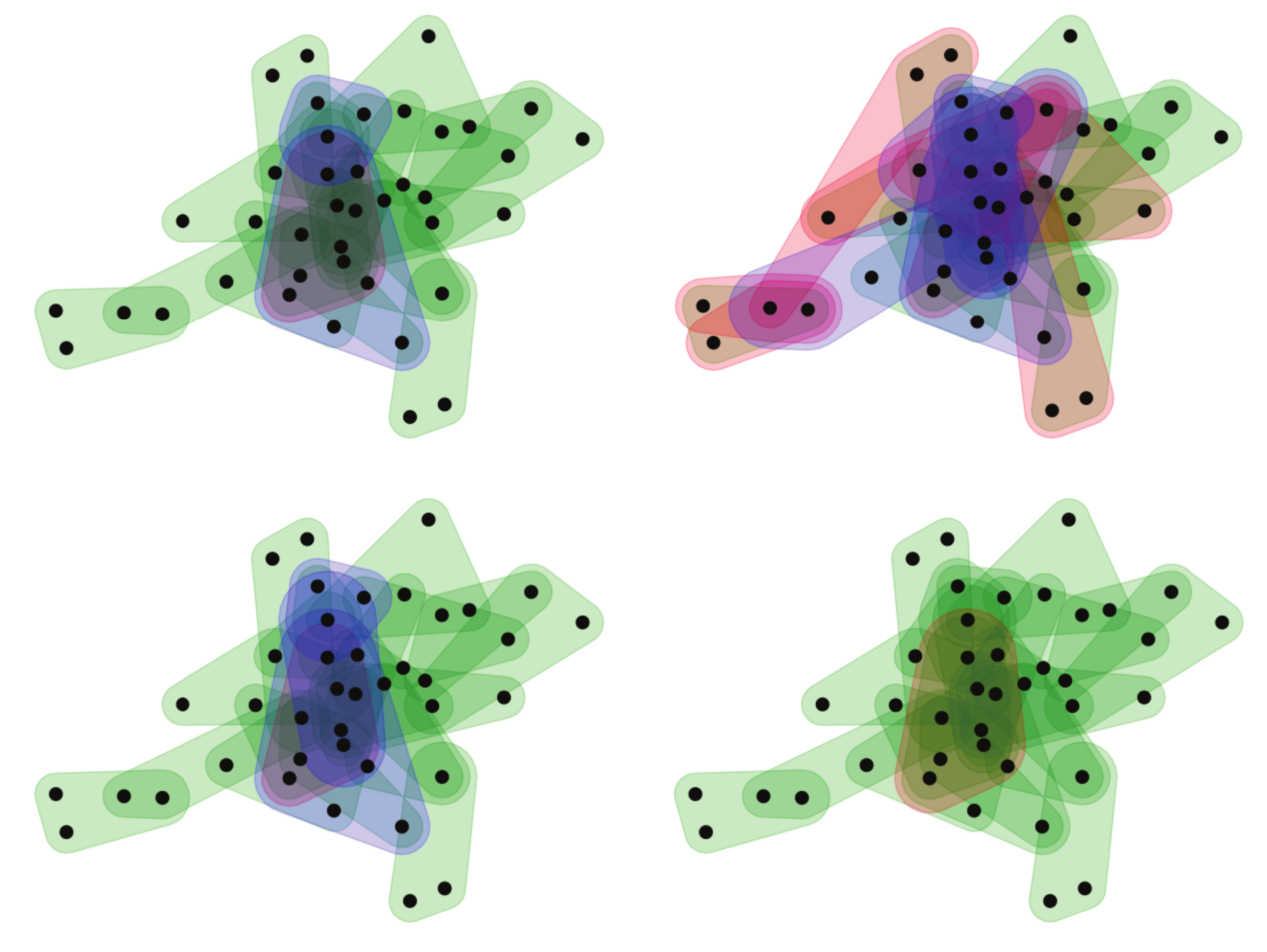}
    \caption{Hypergraph plots of Uniform DBLP experimental results. Each hypergraph was learned with one of the four smoothness terms of interest: (top left) Sum-Square, (top right) Sum-Absolute, (bottom left) Max-Absolute, and (bottom right) Max-Square. Nodes are represented by the black circles and the learned hyperedges are the colored groupings. The hyperedges are color coded where green indicates a correctly learned hyperedge, red is an incorrectly learned hyperedge, and blue is a missing hyperedge that is in the ground truth but not the learned result.}
    \label{fig:PlotsDBLP}
\end{figure}

We used \(\mathbf{z}\), calculated from \(\mathbf{X}\) and the reduced hyperedge set $\hat{E}(\mathcal{H})$, along with Algorithm \ref{alg:HSLS} to get the resultant \(\mathbf{w}\). To refine our choice of \(\alpha\) and \(\beta\), we used a hyperparameter grid search on a log scale. The statistical results of recovering hypergraph structures for both datasets in the uniform and non-uniform scenarios can be found in Table \ref{tab:CoauthorshipResults}. A visual result of the DBLP uniform testing using HSLS can be found in Fig.~\ref{fig:PlotsDBLP}.

\begin{table*}[!t]
   \captionsetup{justification=centering}
   \caption{Co-authorship network experimental metric results.}  
   \label{tab:CoauthorshipResults}
   \centering
   \resizebox{0.7\linewidth}{!}{
   \begin{tabular}{rl|ccc|ccc}
   \toprule\toprule
    & & \multicolumn{3}{c|}{\textbf{Uniform}} & \multicolumn{3}{c}{\textbf{Non-Uniform}}\\
    \cmidrule(lr){3-5}\cmidrule(lr){6-8}
    & \textbf{Learning Method} & \textbf{F1-Score} & \textbf{Precision} & \textbf{Recall} & \textbf{F1-Score} & \textbf{Precision} & \textbf{Recall} \\ 
   \midrule
   
   \multirow{7}{*}{\textbf{DBLP}}                 & PDL-HGSP~\cite{Karelia}                & 0.9130 & 0.8750 & 0.9545 & 0.7143 & 0.5932 & 0.8974 \\
                                                  & Gao et al.~\cite{HGNN+}     & 0.8000 & 0.7143 & 0.9091 & 0.5625 & 0.4045 & \textbf{0.9231} \\
                                                  & HGSI~\cite{Tang}            & 0.9091 & 0.9091 & 0.9091 & 0.4103 & 0.4103 & 0.4103 \\
                                                  & HSLS with Sum-Square        & 0.9302 & 0.9524 & 0.9091 & 0.3133 & 0.2955 & 0.3333 \\
                                                  & HSLS with Sum-Absolute        & 0.7458 & 0.5946 & \textbf{1.0000} & 0.3146 & 0.2800 & 0.3590 \\
                                                  & HSLS with Max-Absolute        & 0.9048 & 0.9500 & 0.8636 & 0.6750 & 0.6585 & 0.6923 \\
                                                  & HSLS with Max-Square          & \textbf{0.9778} & \textbf{0.9565} & \textbf{1.0000} & \textbf{0.8684} & \textbf{0.8919} & 0.8462 \\
   \midrule

   \multirow{7}{*}{\textbf{Cora}}                & PDL-HGSP~\cite{Karelia}   & \textbf{1.0000} & \textbf{1.0000} & \textbf{1.0000} & 0.8000 & 0.7755 & 0.8261 \\
                                                 & Gao et al.~\cite{HGNN+}     & 0.7200 & 0.6429 & 0.8182 & 0.5755 & 0.4396 & \textbf{0.8333} \\
                                                 & HGSI~\cite{Tang}              & 0.8182 & 0.8182 & 0.8182 & 0.4583 & 0.4583 & 0.4583 \\
                                                 & HSLS with Sum-Square          & 0.8421 & 0.8889 & 0.8000 & 0.4946 & 0.5111 & 0.4792 \\
                                                 & HSLS with Sum-Absolute        & 0.9000 & 0.9000 & 0.9000 & 0.4694 & 0.4600 & 0.4792 \\
                                                 & HSLS with Max-Absolute        & 0.9524 & 0.9091 & \textbf{1.0000} & 0.5476 & 0.6389 & 0.4792 \\
                                                 & HSLS with Max-Square          & \textbf{1.0000} & \textbf{1.0000} & \textbf{1.0000} & \textbf{0.8478} & \textbf{0.8864} & 0.8125 \\ 
   \bottomrule
   
   \end{tabular}
   }
\end{table*}

We find that, in the uniform scenarios, our proposed method combined with the Max-Square total variation is always the top performer or equivalent to the best results. In the non-uniform cases, our proposed method with the Max-Square total variation produces the best F1-score and precision, but the K-NN based method of Gao et al. has the best recall. This is to be expected as recall is a measure of how many learned hyperedges are in the final result without regard to incorrect hyperedges. The use of K-NN simply recovered a larger set of hyperedges than all other methods thus increasing the chances of including a hyperedge from the ground truth. However, this is at the sacrifice of precision which explains why the F1-score is dragged down. Generally, the use of K-NN to recover the structure proves to be a reasonable approximation, especially based on the metrics of the uniform scenarios.

HSLS with the other selected total variation terms besides Max-Square performed decent in the uniform cases, often producing better results than the compared methods. However, only the Max-Absolute total variation maintained some of this statistical response in the non-uniform case. Comparatively, HSLS with Max-Square still outperformed the other selected total variations. Since F1-score is the combination of precision and recall, it is the most apt description of overall performance. Therefore, our method was the strongest performer across the board when combined with Max-Square. We also note that all tested methods demonstrate a steep decrease in statistical output in a non-uniform structure application, thus demonstrating the more challenging nature of increasingly complex networks.

\begin{figure*}[t!]
    \centering
    \includegraphics[width=0.9\linewidth]{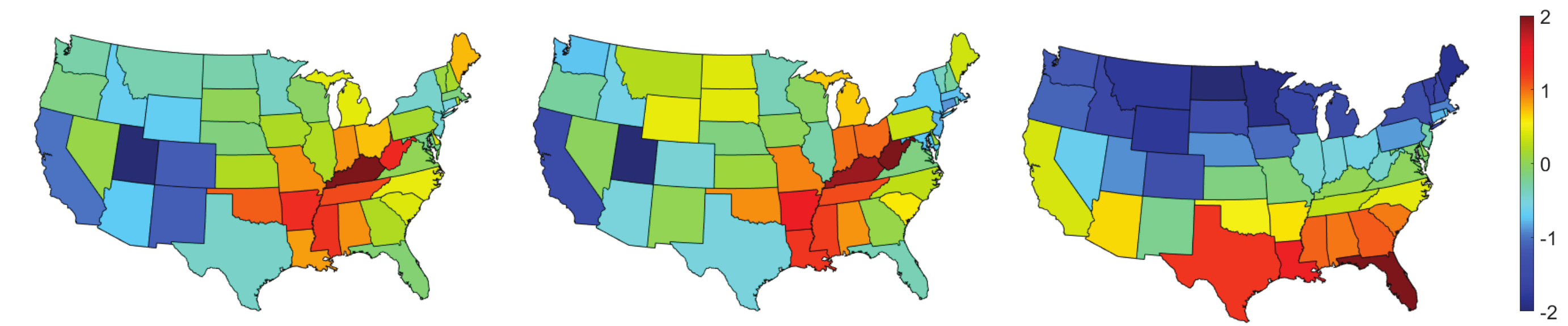}
    \caption{Color maps of data on the contiguous U.S. map. The corresponding datasets are: (left) mortality, (middle) smoking, (right) temperature. All data is averaged across the 2011-2019 observations. The color-bar represents the separately normalized z-score for each dataset.}
    \label{fig:USA}
\end{figure*}

\subsection{Lung Cancer Mortality Rates}
From the co-authorship experiments, the evaluation metrics have demonstrated HSLS can produce reasonable results, especially in uniform cardinality applications. We now evaluate our proposed learning method on real-world data under the assertion that a smooth hypergraph structure recovery is possible with our learning algorithm. We used a dataset reporting lung cancer mortality rates on a year by year basis for each of the contiguous states in the U.S. from 2011-2019 reported in number of people per 100,000~\cite{Lung}. This data is cast as a hypergraph learning problem where each state is a node and the mortality rates are the node signals. By gathering the mortality rates from all 9 years, we create a time series \(\mathbf{X}\) with \(P=9\) signal observations.

The goal was to learn a uniform hypergraph that captures a smooth structure across the nodes to evaluate the emphasis our method puts on hypergraph smoothness. To do this, we first created a baseline hypergraph using K-NN on the mortality data to generate a uniform cardinality hypergraph where \(\mathcal{K}=\{3\}\). K-NN is biased towards creating hyperedges that are locally smooth, but not necessarily globally across the entire hypergraph. However, as we previously investigated with the co-authorship networks, K-NN selected hyperedges can produce reasonable approximations of globally smooth hypergraphs. To that end, we applied each of the four proposed total variations to the baseline K-NN hypergraph structure. Then, we used HSLS to recover hypergraphs with the mortality data. We then calculated the corresponding total variation of our learned hypergraphs for comparison with the baseline. The idea here is that the baseline hypergraph should be relatively smooth, due to K-NN local properties, but our method should encourage better results that are globally smooth. Since the magnitude of the total variation calculations are dependent on the number of hyperedges, we constrain our optimization to have at least as many hyperedges as the baseline.

To further demonstrate the experimental smoothness properties, we introduced two other datasets. The first contains smoking rates \cite{Smoking} and the second contains temperature readings \cite{Temperature}. Both datasets are reported on a year by year basis from 2011-2019 and are therefore of a similar format to the mortality data. We then used the hypergraph structure learned on the mortality data to calculate the corresponding total variation with the smoking and temperature readings. It is well known that the onset of lung cancer and smoking are certainly correlated factors \cite{LungSmokeCorrelation}. Furthermore, the correlation between lung cancer and temperature is at best ambiguous, but certainly not as strong as smoking. To demonstrate this idea, Fig.~\ref{fig:USA} contains color maps of the U.S. based on average lung cancer mortality (Fig.~\ref{fig:USA} left), average smoking rates (Fig.~\ref{fig:USA} middle), and average temperature (Fig.~\ref{fig:USA} right) across the year. The data has been normalized so the figures are on comparable scales. Visually, it is clear that the coloring of the mortality and smoking maps share similar trends, especially in the southeast and northeast regions. These trends become less similar when comparing these two maps to the temperature map which gradients across the U.S. from north to south. So, we expect that if our method can successfully learn a smooth hypergraph on the mortality data, then correlated factors, such as smoking rates, would also be approximately smooth on the same structure. We also expect that factors uncorrelated to the mortality data, such as temperature, should be less smooth across the structure than that of the correlated factors. The results of using K-NN, the learned hypergraphs, and correlated factors are all reported in Table \ref{tab:LungCancerResults}.

\begin{table}[!t]
    \captionsetup{justification=centering}
    \caption{Results of uniform hypergraph learning using the structure recovered from mortality data.}  
    \label{tab:LungCancerResults}
    \centering
    \resizebox{\linewidth}{!}{
    \begin{tabular}{ccccc}
    \toprule\toprule
      & \textbf{Sum-Square} & \textbf{Sum-Absolute} & \textbf{Max-Absolute} & \textbf{Max-Square} \\ 
    \midrule
    \textbf{K-NN TV} & 59.3 & 148.5 & 63.0 & 28.5 \\
    \midrule
    \textbf{K-NN  $|E(\mathcal{H})|$} & 32 & 32 & 32 & 32 \\
    \midrule
    \textbf{HSLS TV} & 41.9 & 142.7 & 60.4 & 11.5 \\
    \midrule
    \textbf{HSLS $|E(\mathcal{H})|$} & 38 & 32 & 32 & 34 \\
    \midrule
    \textbf{Smoking TV} & 305.5 & 431.1 & 206.1 & 162.0 \\
    \midrule
    \textbf{Temperature TV} & 2006.7 & 922.2 & 500.3 & 1184.6 \\
    \bottomrule
   
   \end{tabular}
   }
\end{table}

HSLS, when compared with the baseline K-NN learned hypergraph, always produced smoother structures on the mortality data, even in the presence of more hyperedges. The Sum-Square and Max-Square total variations generate the largest disparity between the baseline and the algorithmic smoothness measure. Furthermore, we note that during the learning process, the sets of hyperedges learned by K-NN and HSLS always had overlap. This makes sense as K-NN promotes local smoothness which can occasionally be optimal for global smoothness as well. The total variation using the smoking data is across the board much greater than that of the baseline and algorithmic approaches. However, comparing the metrics of the smoking and temperature data, we find that the correlated smoking signal was always significantly smoother on the original hypergraph than the less correlated temperature. Therefore, we believe that the learned hypergraph could be used to distinguish between potentially correlated and uncorrelated variables to the initial dataset.

\subsection{Hyperspectral Image Classification}
\begin{figure}[!t]
    \centering
    \subfloat[]{\includegraphics[width=1.4in]{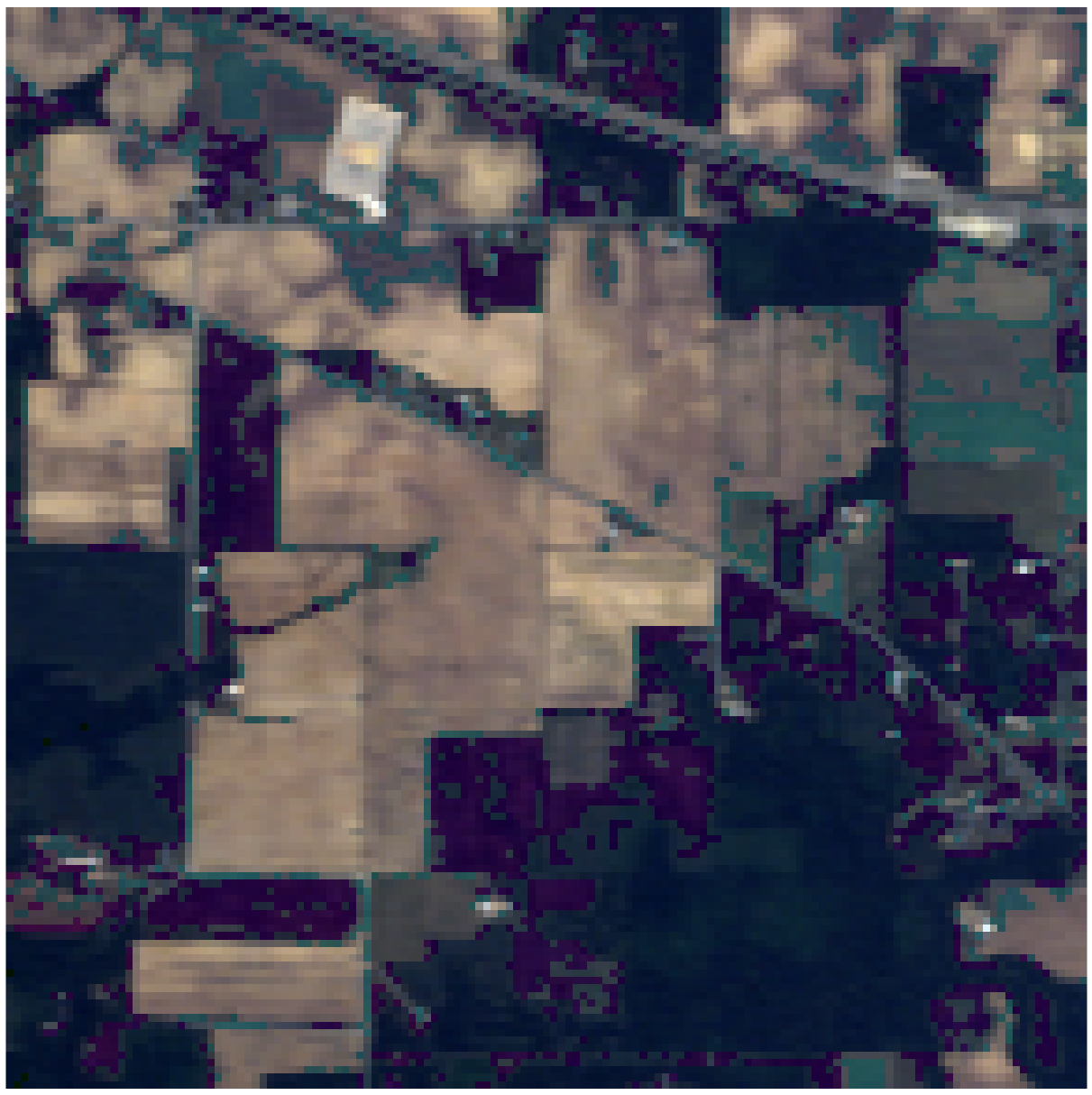}
    \label{fig:FalseColor}}
    \subfloat[]{\includegraphics[width=1.405in]{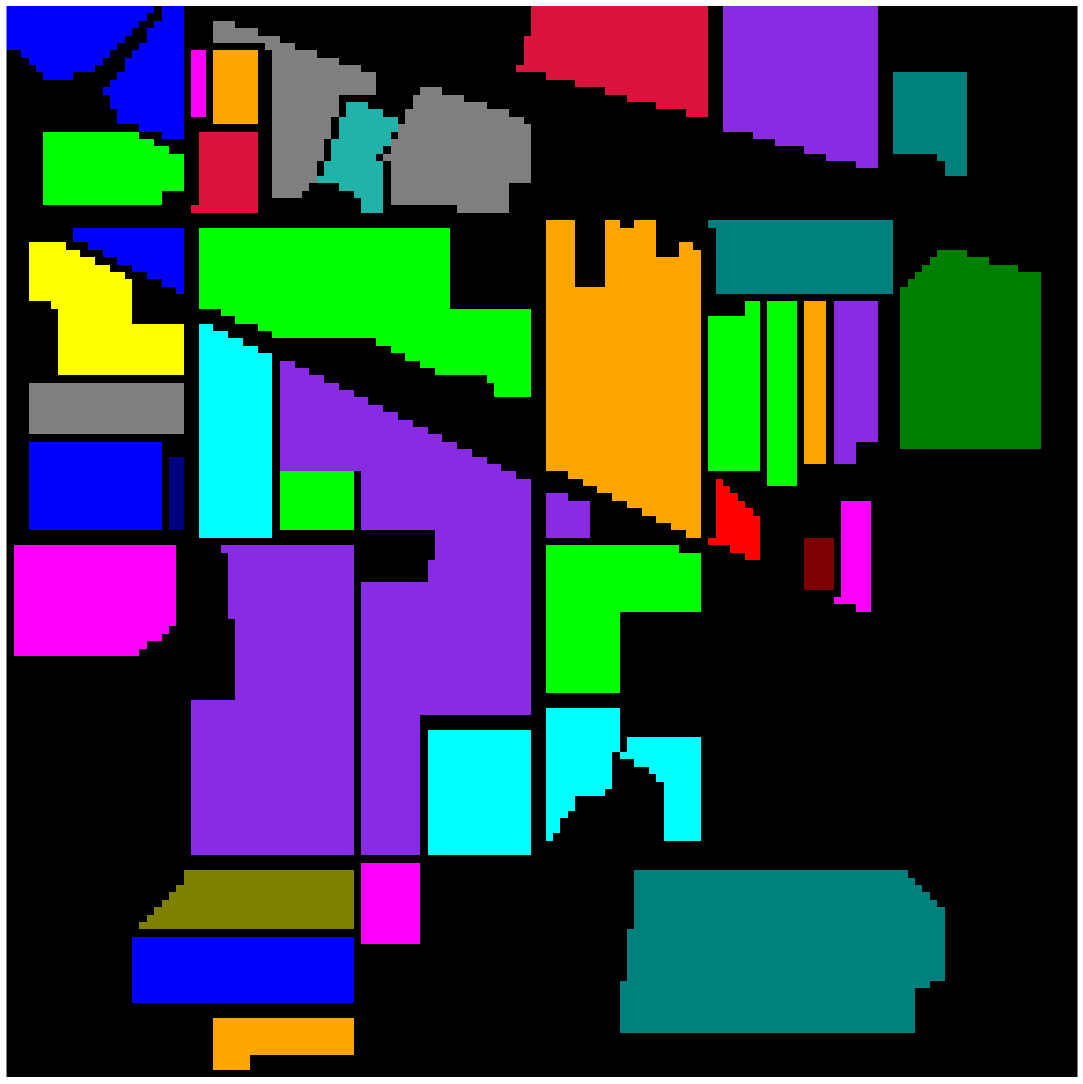}
    \label{fig:GroundTruthHSI}}
    
    \caption{Images based on HSI data: (a) False-color pixel map, (b) Ground truth pixel map with black color for null pixels.}
    \label{fig:HSI}
\end{figure}

A hyperspectral image (HSI) captures spectral information across a range of wavelengths, including those outside of the visible light spectrum. We selected a HSI benchmark dataset known as Indian Pines to perform classification on. This data was originally captured by the Airborne Visible/Infrared Imaging Spectrometer (AVIRIS) sensor over the Purdue University Agronomy farm~\cite{Purdue}, though for this work we use the filtered version of the HSI where 20 bands related to water absorption are removed~\cite{HyperspectralData}. This leaves us with a HSI of 145 x 145 pixels, each with 200 spectral reflectance band features in the 400-2500 nm wavelength range. Figure~\ref{fig:FalseColor} contains a false-color depiction of the HSI being tested on. Each pixel has been assigned a class label 0-16 where label 0 is a null class and will be omitted from here on. The remaining classes 1-16 can be identified by name and quantity in Table~\ref{tab:DataSplits}. Cross referencing the colors of Table~\ref{tab:DataSplits} with Fig.~\ref{fig:GroundTruthHSI} reveals the location of pixel classes in the original HSI.

The goal of this experiment is to use HSLS to improve upon the results of an existing hypergraph based neural network to demonstrate the method's applicability and continue observing the impact of the diverse total variation measurements. The existing neural network framework we chose from Ma et al.~\cite{HyperspectralMethod}, known as $\text{F}^{2}\text{HNN}_{SCC}$, is based on fusion of spectral, spatial, and convolutional neural network (CNN) features. The general framework starts with pre-training a 3-D CNN to generate $\mathbf{X}_{CNN}$ with features for each pixel. Then, the spatial coordinate location of the pixels are used to populate $\mathbf{X}_{spa}$, and the flattened features of the HSI define $\mathbf{X}_{spe}$ the spectral features. From these three generated features sets, corresponding hypergraph incidence matrices $\mathbf{H}_{spe}, \mathbf{H}_{spa}$, and $\mathbf{H}_{CNN}$ are constructed from a radial basis function. The fusion occurs by concatenating $\mathbf{H} = [\mathbf{H}_{spe}, \mathbf{H}_{spa}, \mathbf{H}_{CNN}]$ where $\mathbf{H}$ inputs into a hypergraph convolution layer as
\begin{equation}
    \mathbf{X}^{s+1} = \sigma(\mathbf{D}_v^{-1/2}\mathbf{H}\mathbf{W}\mathbf{D}_e^{-1}\mathbf{H}^\top\mathbf{D}_v^{-1/2}\mathbf{X}^{(s)}\mathbf{\Theta}^{(s)})
\end{equation}
where $\mathbf{X}^{s+1}$ is the output of the $s$-th layer, $\sigma$ is the nonlinear activation, $\mathbf{D}_v$ is the diagonal node degree matrix with diagonal elements $d_{v_i}=\sum_{\mathbf{e}_j \in E(\mathcal{H})} w_{\mathbf{e}_j}\mathbf{H}_{ij}$, $\mathbf{D}_e$ is the diagonal hyperedge degree matrix with diagonal elements $d_{e_j}=\sum_{v_i \in V(\mathcal{H})} \mathbf{H}_{ij}$, and $\mathbf{W}$ is the diagonal hyperedge weight matrix with diagonal elements $\mathbf{w}$. $\text{F}^{2}\text{HNN}_{SCC}$ passes $\mathbf{H}$ through two hypergraph convolution layers, where softmax is then applied row-wise to get the final classification predictions. We refer the readers to~\cite{HyperspectralMethod} for a complete view of the neural network architecture as we implement their work directly.

\begin{table}[!t]
    \captionsetup{justification=centering}
    \caption{Indian Pines Data Splits.}
    \label{tab:DataSplits}
    \centering
    \resizebox{0.85\linewidth}{!}{
    \begin{tabular}{c|c|ccc}
    \toprule
    \textbf{Class} & \textbf{Class Color} & \textbf{Class Name} & \textbf{Training} & \textbf{Testing} \\
    \midrule
    \textbf{1} & \textcolor{myBrightRed}{\rule{0.5cm}{0.2cm}} & Alfalfa & 15 & 31 \\
    \textbf{2} & \textcolor{myBrightGreen}{\rule{0.5cm}{0.2cm}} & Corn Notill & 50 & 1378\\
    \textbf{3} & \textcolor{myBrightBlue}{\rule{0.5cm}{0.2cm}} & Corn Mintill& 50 & 780\\
    \textbf{4} & \textcolor{myYellow}{\rule{0.5cm}{0.2cm}} & Corn & 50 & 187\\
    \textbf{5} & \textcolor{myMagenta}{\rule{0.5cm}{0.2cm}} & Grass Pasture & 50 & 433\\
    \textbf{6} & \textcolor{myCyan}{\rule{0.5cm}{0.2cm}} & Grass Trees & 50 & 680\\
    \textbf{7} & \textcolor{myDarkRed}{\rule{0.5cm}{0.2cm}} & Grass Pasture Mowed & 15 & 13\\
    \textbf{8} & \textcolor{myDarkGreen}{\rule{0.5cm}{0.2cm}} & Hay Windrowed & 50 & 428\\
    \textbf{9} & \textcolor{myDarkBlue}{\rule{0.5cm}{0.2cm}} & Oats & 15 & 5\\
    \textbf{10} & \textcolor{myOrange}{\rule{0.5cm}{0.2cm}} & Soybean Notill & 50 & 922\\
    \textbf{11} & \textcolor{myPurple}{\rule{0.5cm}{0.2cm}} & Soybean Mintill & 50 & 2405\\
    \textbf{12} & \textcolor{myGray}{\rule{0.5cm}{0.2cm}} & Soybean Clean & 50 & 543\\
    \textbf{13} & \textcolor{myOlive}{\rule{0.5cm}{0.2cm}} & Wheat & 50 & 155\\
    \textbf{14} & \textcolor{myTeal}{\rule{0.5cm}{0.2cm}}& Woods & 50 & 1215\\
    \textbf{15} & \textcolor{myCrimson}{\rule{0.5cm}{0.2cm}} & Buildings Grass Trees Drives & 50 & 336\\
    \textbf{16} & \textcolor{myTurquoise}{\rule{0.5cm}{0.2cm}}& Stone Steel Towers & 50 & 43\\
    \midrule
    & & Total & 695 & 9554 \\
    \bottomrule

   \end{tabular}
   }
\end{table}

We note that superpixel segmentation using Simple Linear Iterative Clustering (SLIC) was added as a pre-processing step, which is commonly used in such applications~\cite{SLIC1, SLIC2}. The superpixels are comprised of groups of pixels segmented by SLIC and thereby reduce the spatial dimension. So, for our implementation, the 145 x 145 pixel map was reduced to 194 superpixels. We also update $\mathbf{X}_{spe} \in \mathbb{R}^{194 \times 200}$ and $\mathbf{X}_{CNN} \in \mathbb{R}^{194 \times 256}$ where each row is the average sum of the features from the pixels in each superpixel group. $\mathbf{X}_{spa} \in \mathbb{R}^{194 \times 2}$ now contains the centroid locations of the superpixels.

\begin{table*}[!t]
   \captionsetup{justification=centering}
   \caption{Indian Pines classification results on baseline method $\text{F}^{2}\text{HNN}_{SCC}$ and several implementations of HSLS.}
   \label{tab:ClassificationResults}
   \centering
   \resizebox{\textwidth}{!}{
   \begin{tabular}{c|c|cccccccc}
   \toprule\toprule
   Class & \textbf{$\textbf{F}^{2}\textbf{HNN}_{SCC}$~\cite{HyperspectralMethod}} & \textbf{Uni Sum-Square} & \textbf{Uni Sum-Abs} & \textbf{Uni Max-Abs} & \textbf{Uni Max-Square} & \textbf{Non-Uni Sum-Square} & \textbf{Non-Uni Sum-Abs} & \textbf{Non-Uni Max-Abs} &  \textbf{Non-Uni Max-Square}\\ 
   \midrule
   
   1        & \textbf{97.42 $\pm$ 1.44} & \textbf{97.42 $\pm$ 1.44} & \textbf{97.42 $\pm$ 1.44} & \textbf{97.42 $\pm$ 1.44} & \textbf{97.42 $\pm$ 1.44} & \textbf{97.42 $\pm$ 1.44} & \textbf{97.42 $\pm$ 1.44} & \textbf{97.42 $\pm$ 1.44} & \textbf{97.42 $\pm$ 1.44}  \\
   2        & 69.07 $\pm$ 9.12 & 83.88 $\pm$ 8.15 & 82.31 $\pm$ 10.70 & 83.35 $\pm$ 6.36 & \textbf{86.91 $\pm$ 5.24} & 81.28 $\pm$ 9.31 & 85.54 $\pm$ 5.51 & 80.46 $\pm$ 5.20 & 84.80 $\pm$ 8.04  \\
   3        & 77.82 $\pm$ 26.19 & 88.08 $\pm$ 5.53 & 78.13 $\pm$ 14.26 & 81.44 $\pm$ 8.41 & 86.49 $\pm$ 7.79 & 88.54 $\pm$ 5.68 & 86.56 $\pm$ 7.69 & 62.33 $\pm$ 10.33 & \textbf{88.95 $\pm$ 3.49}  \\
   4        & 94.33 $\pm$ 2.66 & 93.69 $\pm$ 1.33 & 93.05 $\pm$ 0.76 & 93.05 $\pm$ 0.76 & 93.05 $\pm$ 0.76 & 94.01 $\pm$ 1.62 & 93.37 $\pm$ 1.39 & \textbf{96.26 $\pm$ 0.76} & 93.69 $\pm$ 1.33  \\
   5        & 69.61 $\pm$ 24.01 & 87.76 $\pm$ 5.97 & 86.56 $\pm$ 4.19 & 88.73 $\pm$ 5.30 & \textbf{92.06 $\pm$ 4.56} & 88.82 $\pm$ 4.52 & 88.73 $\pm$ 6.19 & 89.61 $\pm$ 6.02 & 87.67 $\pm$ 6.83  \\
   6        & \textbf{93.56} $\pm$ 4.15 & 89.15 $\pm$ 4.06 & 89.09 $\pm$ 4.04 & 89.21 $\pm$ 4.10 & 89.09 $\pm$ 4.04 & 89.03 $\pm$ 4.00 & 89.03 $\pm$ 4.00 & 84.35 $\pm$ 7.56 & 89.03 $\pm$ 4.00  \\
   7        & \textbf{92.31 $\pm$ 17.20} & 84.62 $\pm$ 21.07 & \textbf{92.31 $\pm$ 17.20} & 78.46 $\pm$ 19.91 & 70.77 $\pm$ 16.68 & 84.62 $\pm$ 21.07 & 76.92 $\pm$ 21.07 & 66.15 $\pm$ 40.92 & 84.62 $\pm$ 21.07  \\
   8        & 99.21 $\pm$ 1.09 & 99.16 $\pm$ 1.15 & 99.21 $\pm$ 1.09 & \textbf{99.58 $\pm$ 0.94} & 99.16 $\pm$ 1.15 & \textbf{99.58 $\pm$ 0.94} & \textbf{99.58 $\pm$ 0.94} & 98.36 $\pm$ 0.92 & 99.16 $\pm$ 1.15  \\
   9        & 60.00 $\pm$ 54.77 & \textbf{100.00 $\pm$ 0.00} & \textbf{100.00 $\pm$ 0.00} & \textbf{100.00 $\pm$ 0.00} & \textbf{100.00 $\pm$ 0.00} & \textbf{100.00 $\pm$ 0.00} & \textbf{100.00 $\pm$ 0.00} & 72.00 $\pm$ 38.99 & \textbf{100.00 $\pm$ 0.00}  \\
   10       & 68.16 $\pm$ 25.29 & 78.31 $\pm$ 5.20 & 79.48 $\pm$ 9.66 & 79.57 $\pm$ 4.72 & 75.29 $\pm$ 8.88 & 76.62 $\pm$ 11.24 & 79.11 $\pm$ 1.72 & \textbf{80.98 $\pm$ 3.27} & 76.36 $\pm$ 8.77  \\
   11       & 68.30 $\pm$ 20.79 & 87.44 $\pm$ 5.62 & 90.38 $\pm$ 5.15 & 88.42 $\pm$ 6.54 & 87.64 $\pm$ 4.02 & 88.62 $\pm$ 3.87 & \textbf{90.41 $\pm$ 3.92} & 78.55 $\pm$ 14.12 & 87.81 $\pm$6.18  \\
   12       & 78.97 $\pm$ 22.88 & 86.70 $\pm$ 6.14 & \textbf{88.77 $\pm$ 3.89} & \textbf{88.77 $\pm$ 3.89} & 87.74 $\pm$ 3.45 & 83.20 $\pm$ 7.58 & 87.04 $\pm$ 6.28 & 76.21 $\pm$ 3.13 & 86.23 $\pm$ 5.79  \\
   13       & 98.71 $\pm$ 0.91 & 98.71 $\pm$ 0.91 & 98.71 $\pm$ 0.91 & 98.71 $\pm$ 0.91 & 98.71 $\pm$ 0.91 & 98.71 $\pm$ 0.91 & 98.71 $\pm$ 0.91 & \textbf{99.23 $\pm$ 0.71} & 98.71 $\pm$ 0.91  \\
   14       & 97.70 $\pm$ 2.91 & 98.42 $\pm$ 2.49 & \textbf{99.56 $\pm$ 0.99} & \textbf{99.56 $\pm$ 0.99} & 98.42 $\pm$ 2.49 & 99.54 $\pm$ 0.99 & 99.54 $\pm$ 0.99 & 94.03 $\pm$ 0.99 & 99.54 $\pm$ 0.99  \\
   15       & 98.04 $\pm$ 0.27 & 98.75 $\pm$ 1.14 & 98.45 $\pm$ 0.90 & 98.04 $\pm$ 0.27 & 98.75 $\pm$ 1.14 & 93.81 $\pm$ 9.52 & 98.45 $\pm$ 0.90 & 98.33 $\pm$ 0.93 & \textbf{99.17 $\pm$ 1.14}  \\
   16       & 79.07 $\pm$ 44.22 & \textbf{98.61 $\pm$ 1.27} & \textbf{98.61 $\pm$ 1.27} & \textbf{98.61 $\pm$ 1.27} & \textbf{98.61 $\pm$ 1.27} & \textbf{98.61 $\pm$ 1.27} & \textbf{98.61 $\pm$ 1.27} & \textbf{98.61 $\pm$ 1.27} & \textbf{98.61 $\pm$ 1.27}  \\
   \midrule
   OA (\%)   & 78.98 $\pm$ 8.97 & 88.90 $\pm$ 0.87 & 88.91 $\pm$ 1.72 & 88.93 $\pm$ 1.17 & 89.19 $\pm$ 0.69 & 88.53 $\pm$ 3.06 & \textbf{90.03 $\pm$ 0.36} & 82.88 $\pm$ 3.95 & 89.13 $\pm$ 0.82 \\
   AA (\%)   & 83.89 $\pm$ 10.85 & 91.92 $\pm$ 1.01 & \textbf{92.00 $\pm$ 1.24} & 91.43 $\pm$ 1.22 & 91.26 $\pm$ 1.43 & 91.40 $\pm$ 1.40 & 91.81 $\pm$ 1.06 & 85.81 $\pm$ 2.27 & 91.98 $\pm$ 1.21  \\
   $\kappa$  & 0.762 $\pm$ 0.10 & 0.873 $\pm$ 0.01 & 0.873 $\pm$ 0.02 & 0.874 $\pm$ 0.01 & 0.876 $\pm$ 0.01 & 0.869 $\pm$ 0.04 & \textbf{0.886 $\pm$ 0.00} & 0.806 $\pm$ 0.04 & 0.876 $\pm$ 0.01 \\
   \bottomrule
   
   \end{tabular}
   }
\end{table*}

We implement $\text{F}^{2}\text{HNN}_{SCC}$ as the baseline method with the recommended radial basis function to generate weighted $\mathbf{H}$. This is compared against the use of HSLS to generate binary $\mathbf{H}$ with $\mathbf{X} = [\mathbf{X}_{spe},\mathbf{X}_{spa},\mathbf{X}_{CNN}]$ being the signal input for both methods. For HSLS, we recover both a uniform hypergraph structure with $\mathcal{K} = \{3\}$ and a non-uniform hypergraph structure with $\mathcal{K} = \{2,3,4\}$. All proposed total variation terms are tested to recover $\mathbf{w}$, which in turn was used to populate the binary $\mathbf{H}$, and these terms together can generate $\mathbf{D}_v$, $\mathbf{D}_e$, and $\mathbf{W}$. Algorithm~\ref{alg:hyperedge_reduction} was used on $\mathbf{X}$ to reduce the hyperedge search space, creating a scalable problem and enabling the feasibility of the non-uniform testing. After generating $\mathbf{H}$, the hypergraph convolutional neural network was applied which outputs classification predictions.

Both the baseline and HSLS methods have the same training and testing scheme and were evaluated on the same training and testing splits. We note that the selected training method is derivative of other works in HSI classification~\cite{HyperspectralMethod,GCNHSI}. Table~\ref{tab:DataSplits} reports the number of training and testing samples per class. We implemented a Monte Carlo cross-validation setup such that, for 5 total passes, the training and testing splits were randomized, but adhered to Table~\ref{tab:DataSplits}. Then, 4 validation samples per class were randomly sampled from the training data. The neural network model was trained over 300 epochs using the Adam optimizer and cross entropy loss, starting at a learning rate of 0.01 and decreasing based on a learning rate scheduler. Every epoch, the model was evaluated on the validation set and the validation loss was recorded. For more details, we refer the reader to the code provided. After training, the model with the minimum validation loss was used for testing. The test results were averaged across 5 trials and are reported in Table~\ref{tab:ClassificationResults} where the baseline is $\text{F}^{2}\text{HNN}_{SCC}$, and all other columns are HSLS either uniform (written as Uni) or non-uniform (written as Non-Uni) combined with one of the four total variation terms (with Absolute written as Abs). The majority of the table reports accuracy per class, OA is the overall accuracy, AA is the average accuracy found from the mean of the sum of individual class accuracies, and $\kappa$ is Cohen's kappa value, a measurement of how closely the model agrees with the ground truth labels while accounting for random chance.

The use of HSLS across all variants always improves upon the OA, AA, and $\kappa$ of the baseline value from $\text{F}^{2}\text{HNN}_{SCC}$. Furthermore, the per class accuracy is also generally improved, with few exceptions. This justifies the use of HSLS as a viable alternative for hypergraph structure inference for use in downstream tasks. Comparing the results among the HSLS models, the best performer could be attributed to the non-uniform Sum-Absolute implementation, though many of the overall results are within a percent or two of each other, with the worst result coming from the non-uniform Max-Absolute model. This is a striking difference to the results of the co-authorship networks, as all total variation terms seem to produce viable outputs for classification. This could be contributed to the fluidity of the HSI structure where, in the absence of a ground truth hypergraph, there are many hypergraphs that could appropriately describe some underlying topology. This supports the conclusion that the effectiveness of hypergraph total variation terms are application dependent.

%% file: Sections/conclusion.tex
\section{Conclusion} \label{sec:Conclusion}

In this paper, we developed a novel interpretation of a hypergraph learning optimization method, HSLS, which aimed to remedy the lack of a smoothness prior, shortage of total variation testing, algorithmic assumptions, and scalability. We overcame the first two problems by basing our optimization framework on minimizing the total variation while allowing the smoothness term to be robust for substitution. A diverse set of smoothness terms were introduced such that the efficacy of a range of definitions could be tested in different applications. The choice to format our optimization with a primal and dual variable and the investigation into the use of K-NN as a means of hyperedge possibility reduction solved the issues surrounding the algorithmic assumptions related to weighted hyperedge importance and valid hyperedge search spaces. Furthermore, the K-NN reduction method we improved upon made our method scalable without compromising the span of the hyperedge search. We were able to experimentally evaluate the effectiveness of our proposed method, with each selected total variation, in recovering a hypergraph structure and thereby generated original empirical results.

When investigating our method's ability to accurately recover a hypergraph with reference to the ground truth topology, we found that our proposed method HSLS with Max-Square was the overall best performer when compared to the other smoothness terms and the results of other methods. This validated our optimization framework showing that reasonable hypergraph structure recovery was possible and total variation substitution was viable. Furthermore, this supported the conclusion that Max-Square, of the tested total variations, would be the best choice of smoothness in the application of recovering a specific hypergraph structure.

We proceeded with evaluating how well our method encouraged smooth hypergraph structures. We found that, across all total variations, our method encouraged global smoothness more so than the K-NN baseline. We also determined that our method recovers a structure that is smoother across data correlated to the original dataset and less smooth across uncorrelated data, highlighting a key property of the learned hypergraph structure.

Finally, we demonstrated our method's ability to scale to larger applications, defined by more nodes and higher cardinalities, through improving upon an existing hypergraph based method of hyperspectral image classification. The metric report demonstrated the performance increase of using HSLS over the baseline and revealed that, in a domain where the ground truth hypergraph is not present, all total variation terms across uniform and non-uniform scenarios were viable candidates for better results, which is similar to the outcome of the mortality rate experimentation.

In future work, we will extend concepts to a dynamic domain where time-series data is subject to alter as more observations are taken. This implies that the hypergraph structure at one point in time may not be consistent with future points in time. We believe that our work here is a suitable starting point to evolve the research in this direction.

%% file: Sections/acknowledgement.tex
\section*{Acknowledgement}
The authors would like to thank Mihir Malladi for their contribution to the work in regard to some of the experimental data acquisition and interpretation.

%% file: Sections/appendix.tex
\appendices
\section{Proof of Equation~\eqref{eq:proxf}} \label{sec:AppendixA}

First, we write out the definition of the proximal operator as
\begin{equation} \label{eq:proxdefinition}
    \text{prox}_{\gamma f}(\mathbf{x}) := \underset{\mathbf{u}}{\text{argmin}} \ f(\mathbf{u})+ \frac{1}{2\gamma} ||\mathbf{u} - \mathbf{x}||_2^2.
\end{equation}

Using the proximal affine addition operation, we establish that if you have the problem of the form \(\hat{f}(\mathbf{x}) = \Psi(\mathbf{x}) + \mathbf{a}^\top \mathbf{x} + \mathbf{b}\), then
\begin{equation} \label{eq:affineaddition}
    \text{prox}_{\gamma \hat{f}}(\mathbf{x}) = \text{prox}_{\gamma \Psi}(\mathbf{x} - \gamma\mathbf{a})
\end{equation}
for some function \(\Psi(\mathbf{x})\) and constant vectors \(\mathbf{a}, \mathbf{b}\). We can directly apply the original function \(f(\mathbf{x}) = \mathbf{z}^\top \mathbf{x} + \mathbbm{1}\{\mathbf{x} \geq 0\}\) where \(\Psi(\mathbf{x}) = \mathbbm{1}\{\mathbf{x} \geq 0\}\), \(\mathbf{a} = \mathbf{z}\), and \(\mathbf{b} = \mathbf{0}\). Using the proximal rule of \eqref{eq:affineaddition}, we can write

\begin{equation} \label{eq:subinaffine}
    \text{prox}_{\gamma f}(\mathbf{x}) = \text{prox}_{\gamma \mathbbm{1}\{\cdot\}}(\mathbf{x} - \gamma \mathbf{z}),
\end{equation}
and substituting into the proximal operator definition of \eqref{eq:proxdefinition} we arrive at
\begin{equation} \label{eq:indicatorprox}
    \begin{aligned}
        \text{prox}_{\gamma \mathbbm{1}\{\cdot\}}&(\mathbf{x} - \gamma \mathbf{z}) = \\
        & \underset{\mathbf{u}}{\text{argmin}} \ \mathbbm{1}\{\mathbf{u} \geq 0\} + \frac{1}{2\gamma} ||\mathbf{u} - (\mathbf{x} - \gamma \mathbf{z})||_2^2.
    \end{aligned}
\end{equation}

With the indicator function, we note that the argument \(\mathbf{u} \geq 0\) is the definition of the indicator function set \(C\) meaning that if the argument satisfies the condition, then that argument is in set \(C\). Since the proximal operator is being performed on the indicator function, this is considered the projection of \(\mathbf{u}\) onto \(C\) which allows the rewrite of \eqref{eq:indicatorprox} as
\begin{equation}
    \text{prox}_{\gamma \mathbbm{1}\{\cdot\}}(\mathbf{x} - \gamma \mathbf{z}) = \underset{\mathbf{u} \in C}{\text{argmin}} \ ||\mathbf{u} - \mathbf{x} + \gamma \mathbf{z}||_2^2
\end{equation}
where we also discarded the \(1/{2\gamma}\) constant that no longer affects the result. We now have the constraint that \(\mathbf{u} \in C\). We proceed with minimizing via the gradient as
\begin{equation} \label{eq:proxfgradient}
    \nabla ||\mathbf{u} - \mathbf{x} + \gamma \mathbf{z}||_2^2 = 2(\mathbf{u} - \mathbf{x} + \gamma \mathbf{z}).
\end{equation}
Then, setting \eqref{eq:proxfgradient} to \(\mathbf{0}\) the result is
\begin{equation}
    \mathbf{u} = \mathbf{x} - \gamma\mathbf{z}.
\end{equation}
The final step is to include the constraint \(\mathbf{u} \in C\). We know that \(u_i\) remains unchanged if it is negative, otherwise it is set to \(+\infty\). So, we can threshold out the negative values and arrive at
\begin{equation}
    (\text{prox}_{\gamma \mathbbm{1}\{\cdot\}}(\mathbf{x} - \gamma \mathbf{z}))_i = \text{max}(0, x_i - \gamma z_i),
\end{equation}
which, when substituting using \eqref{eq:subinaffine}, provides exactly the answer in \eqref{eq:proxf}.

\section{Proof of Equation~\eqref{eq:proxg}}\label{sec:AppendixB}

We first introduce the general form of the Moreau Decomposition as
\begin{equation} \label{eq:Moreau}
     \mathbf{x} = \text{prox}_{\gamma g}(\mathbf{x}) + \gamma\text{prox}_{g^* / \gamma }(\mathbf{x} / \gamma ).
\end{equation}
Using function \(g(\mathbf{x}) = -\alpha \mathbf{1}^\top \log (\mathbf{x})\), it is well established for the logarithmic barrier that
\begin{equation} \label{eq:proxlog}
    (\text{prox}_{\gamma g} (\mathbf{x}))_i = \frac{x_i + \sqrt{x_i^2 + 4\alpha\gamma }}{2},
\end{equation}
which is written in its element-wise form for \(i = 1,2, \dots ,n\) where \(\mathbf{x} \in \mathbb{R}^n\). Here, we establish that the logarithmic barrier \(g\) is convex and closed, and as such the conjugate of the conjugate function produces the original function \((g^*)^* = g\). With this in mind, we can rewrite \eqref{eq:Moreau} as
\begin{equation} \label{eq:Moreaurewrite}
    \text{prox}_{\gamma g}(\mathbf{x}) =  \mathbf{x} - \gamma\text{prox}_{g^* / \gamma }(\mathbf{x} / \gamma ).
\end{equation}
Rewriting the functions to be their conjugates and substituting in \eqref{eq:proxlog}, we get
\begin{align}
    \label{eq:proxgderivation}
    (\text{prox}_{\gamma g^*}(\mathbf{x}))_i &=  x_i - \gamma(\text{prox}_{g / \gamma }(\mathbf{x} / \gamma ))_i \\
    & = x_i - \gamma\left(\frac{x_i / \gamma + \sqrt{(x_i / \gamma)^2 + 4\alpha/\gamma}}{2}\right) \\
    & = x_i - \gamma\left(\frac{y_i + \sqrt{y_i^2 + 4\alpha/\gamma}}{2}\right)
\end{align}
where, after replacing \(x_i / \gamma = y_i\), we get exactly \eqref{eq:proxg}.

\section{Derivation of Learning Rate Sequence \(\gamma\)} \label{sec:AppendixC}

The learning rate sequence \(\gamma\) is dependent on the Lipschitz constant \(\zeta\) of \(h\) from \eqref{eq:convex}. This constant is the minimum value of
\begin{equation} \label{eq:lip}
    ||\nabla h(\mathbf{x}) - \nabla h(\mathbf{y})|| \leq \zeta ||\mathbf{x} - \mathbf{y}||
\end{equation}
where \(\zeta \in \mathbb{R}^+\). Substituting \(\nabla h\) from \eqref{eq:gradh}, we can rewrite \eqref{eq:lip} as
\begin{equation} \label{eq:rewrite_lip}
    ||2\beta\mathbf{x} - 2\beta\mathbf{y}|| \leq \zeta ||\mathbf{x} - \mathbf{y}||.
\end{equation}
Then, we pull out the constant \(2\beta\) and divide both sides by \(||\mathbf{x} - \mathbf{y}||\) to get \(2\beta \leq \zeta\). Since the Lipschitz constant is supposed to be the minimum value, then \(\zeta = 2\beta\).

From here, there is no longer any derivation, only the definition of the sequence of \(\gamma\) defined by
\begin{align}
    \mu &= \zeta + ||S||_s, \\
    \epsilon &\in (0,1/(1+\mu)), \\
    (\gamma_n)_{n \in \mathbb{N}} &\in [\epsilon, (1-\epsilon)/\mu]
\end{align}
where \(||\cdot||_s\) is the spectral norm, \(\epsilon\) is some scalar, and each \(\gamma\) in the sequence is indexed by \(n\).

%% file: main.bbl
\begin{thebibliography}{10}
\providecommand{\url}[1]{#1}
\csname url@samestyle\endcsname
\providecommand{\newblock}{\relax}
\providecommand{\bibinfo}[2]{#2}
\providecommand{\BIBentrySTDinterwordspacing}{\spaceskip=0pt\relax}
\providecommand{\BIBentryALTinterwordstretchfactor}{4}
\providecommand{\BIBentryALTinterwordspacing}{\spaceskip=\fontdimen2\font plus
\BIBentryALTinterwordstretchfactor\fontdimen3\font minus \fontdimen4\font\relax}
\providecommand{\BIBforeignlanguage}[2]{{%
\expandafter\ifx\csname l@#1\endcsname\relax
\typeout{** WARNING: IEEEtran.bst: No hyphenation pattern has been}%
\typeout{** loaded for the language `#1'. Using the pattern for}%
\typeout{** the default language instead.}%
\else
\language=\csname l@#1\endcsname
\fi
#2}}
\providecommand{\BIBdecl}{\relax}
\BIBdecl

\bibitem{HypergraphIntro}
Y.~Gao, Z.~Zhang, H.~Lin, X.~Zhao, S.~Du, and C.~Zou, ``Hypergraph learning: Methods and practices,'' \emph{IEEE Transactions on Pattern Analysis and Machine Intelligence}, vol.~44, no.~5, pp. 2548--2566, 2022.

\bibitem{Email}
\BIBentryALTinterwordspacing
I.~Amburg, J.~Kleinberg, and A.~R. Benson, ``Planted hitting set recovery in hypergraphs,'' \emph{Journal of Physics: Complexity}, vol.~2, no.~3, p. 035004, may 2021. [Online]. Available: \url{https://dx.doi.org/10.1088/2632-072X/abdb7d}
\BIBentrySTDinterwordspacing

\bibitem{Power}
Z.~Liu, C.~Luo, J.~Xie, and Y.~Luo, ``A synchronous training hypergraph neural network for power allocation in multi-cell multi-user networks,'' \emph{IEEE Wireless Communications Letters}, vol.~13, no.~4, pp. 1113--1117, 2024.

\bibitem{Fault}
H.~Ke, Z.~Chen, J.~Xu, X.~Fan, C.~Yang, and T.~Peng, ``Time-frequency hypergraph neural network for rotating machinery fault diagnosis with limited data,'' in \emph{2023 IEEE 12th Data Driven Control and Learning Systems Conference (DDCLS)}, 2023, pp. 1786--1792.

\bibitem{Text}
Y.~Zhang, M.~Guo, Q.~Yan, and G.~Shen, ``Short text classification via hypergraph convolution network,'' in \emph{2021 3rd International Symposium on Smart and Healthy Cities (ISHC)}, 2021, pp. 72--76.

\bibitem{Traffic}
K.~Wang, J.~Chen, S.~Liao, J.~Hou, and Q.~Xiong, ``Geographic-semantic-temporal hypergraph convolutional network for traffic flow prediction,'' in \emph{2020 25th International Conference on Pattern Recognition (ICPR)}, 2021, pp. 5444--5450.

\bibitem{Brain}
A.~N. Pisarchik, N.~P. Serrano, and R.~Jaimes-Reátegui, ``Brain connectivity hypergraphs,'' in \emph{2024 8th Scientific School Dynamics of Complex Networks and their Applications (DCNA)}, 2024, pp. 190--193.

\bibitem{Image}
S.~Zhang, S.~Cui, and Z.~Ding, ``Hypergraph-based image processing,'' in \emph{2020 IEEE International Conference on Image Processing (ICIP)}, 2020, pp. 216--220.

\bibitem{Object}
\BIBentryALTinterwordspacing
L.~Nong, J.~Wang, J.~Lin, H.~Qiu, L.~Zheng, and W.~Zhang, ``Hypergraph wavelet neural networks for {3D} object classification,'' \emph{Neurocomputing}, vol. 463, pp. 580--595, 2021. [Online]. Available: \url{https://www.sciencedirect.com/science/article/pii/S0925231221011905}
\BIBentrySTDinterwordspacing

\bibitem{Segmentation}
J.~H. Giraldo, V.~Scarrica, A.~Staiano, F.~Camastra, and T.~Bouwmans, ``Hypergraph convolutional networks for weakly-supervised semantic segmentation,'' in \emph{2022 IEEE International Conference on Image Processing (ICIP)}, 2022, pp. 16--20.

\bibitem{Noise}
S.~Rital and H.~Cherifi, ``Similarity hypergraph representation for impulsive noise reduction,'' in \emph{Proceedings EC-VIP-MC 2003. 4th EURASIP Conference focused on Video/Image Processing and Multimedia Communications (IEEE Cat. No.03EX667)}, vol.~2, 2003, pp. 539--544 vol.2.

\bibitem{BrainNetworks}
\BIBentryALTinterwordspacing
O.~Sporns, ``Structure and function of complex brain networks,'' \emph{Dialogues in Clinical Neuroscience}, vol.~15, no.~3, pp. 247--262, 2013, pMID: 24174898. [Online]. Available: \url{https://doi.org/10.31887/DCNS.2013.15.3/osporns}
\BIBentrySTDinterwordspacing

\bibitem{SocialNetworks}
\BIBentryALTinterwordspacing
G.~Mateos, S.~Segarra, and A.~Marques, ``Chapter 13 - inference of graph topology,'' in \emph{Cooperative and Graph Signal Processing}, P.~M. Djurić and C.~Richard, Eds.\hskip 1em plus 0.5em minus 0.4em\relax Academic Press, 2018, pp. 349--374. [Online]. Available: \url{https://www.sciencedirect.com/science/article/pii/B9780128136775000134}
\BIBentrySTDinterwordspacing

\bibitem{GraphDiffusion}
D.~Thanou, X.~Dong, D.~Kressner, and P.~Frossard, ``Learning heat diffusion graphs,'' \emph{IEEE Transactions on Signal and Information Processing over Networks}, vol.~3, no.~3, pp. 484--499, 2017.

\bibitem{Dong}
X.~Dong, D.~Thanou, P.~Frossard, and P.~Vandergheynst, ``Learning {Laplacian} matrix in smooth graph signal representations,'' \emph{IEEE Transactions on Signal Processing}, vol.~64, no.~23, pp. 6160--6173, 2016.

\bibitem{Kalofolias}
\BIBentryALTinterwordspacing
V.~Kalofolias, ``How to learn a graph from smooth signals,'' in \emph{Proceedings of the 19th International Conference on Artificial Intelligence and Statistics}, ser. Proceedings of Machine Learning Research, A.~Gretton and C.~C. Robert, Eds., vol.~51.\hskip 1em plus 0.5em minus 0.4em\relax Cadiz, Spain: PMLR, 09--11 May 2016, pp. 920--929. [Online]. Available: \url{https://proceedings.mlr.press/v51/kalofolias16.html}
\BIBentrySTDinterwordspacing

\bibitem{DongKaloCite1}
H.~E. Egilmez, E.~Pavez, and A.~Ortega, ``Graph learning from data under {Laplacian} and structural constraints,'' \emph{IEEE Journal of Selected Topics in Signal Processing}, vol.~11, no.~6, pp. 825--841, 2017.

\bibitem{DongKaloCite2}
\BIBentryALTinterwordspacing
R.~Balestriero and Y.~LeCun, ``Contrastive and non-contrastive self-supervised learning recover global and local spectral embedding methods,'' in \emph{Advances in Neural Information Processing Systems}, S.~Koyejo, S.~Mohamed, A.~Agarwal, D.~Belgrave, K.~Cho, and A.~Oh, Eds., vol.~35.\hskip 1em plus 0.5em minus 0.4em\relax Curran Associates, Inc., 2022, pp. 26\,671--26\,685. [Online]. Available: \url{https://proceedings.neurips.cc/paper_files/paper/2022/file/aa56c74513a5e35768a11f4e82dd7ffb-Paper-Conference.pdf}
\BIBentrySTDinterwordspacing

\bibitem{DongKaloCite3}
S.~Segarra, A.~G. Marques, G.~Mateos, and A.~Ribeiro, ``Network topology inference from spectral templates,'' \emph{IEEE Transactions on Signal and Information Processing over Networks}, vol.~3, no.~3, pp. 467--483, 2017.

\bibitem{MoreGraphSmoothness}
\BIBentryALTinterwordspacing
S.~Chepuri, S.~Liu, G.~Leus, and A.~Hero, ``\BIBforeignlanguage{English}{Learning sparse graphs under smoothness prior},'' in \emph{\BIBforeignlanguage{English}{2017 IEEE International Conference on Acoustics, Speech, and Signal Processing - Proceedings}}.\hskip 1em plus 0.5em minus 0.4em\relax United States: IEEE, 2017, pp. 6508--6512, iCASSP 2017 : 42nd IEEE International Conference on Acoustics, Speech and Signal Processing - The Internet of Signals, ICASSP ; Conference date: 05-03-2017 Through 09-03-2017. [Online]. Available: \url{http://www.ieee-icassp2017.org/}
\BIBentrySTDinterwordspacing

\bibitem{MoreGraphSmoothness2}
\BIBentryALTinterwordspacing
S.~S. Saboksayr, G.~Mateos, and M.~Cetin, ``Online graph learning under smoothness priors,'' \emph{2021 29th European Signal Processing Conference (EUSIPCO)}, 2021. [Online]. Available: \url{https://par.nsf.gov/biblio/10321463}
\BIBentrySTDinterwordspacing

\bibitem{MoreGraphSmoothness3}
\BIBentryALTinterwordspacing
V.~Kalofolias and N.~Perraudin, ``Large scale graph learning from smooth signals,'' in \emph{International Conference on Learning Representations}, 2019. [Online]. Available: \url{https://openreview.net/forum?id=ryGkSo0qYm}
\BIBentrySTDinterwordspacing

\bibitem{MoreSmoothness4}
M.~G. Rabbat, ``Inferring sparse graphs from smooth signals with theoretical guarantees,'' in \emph{2017 IEEE International Conference on Acoustics, Speech and Signal Processing (ICASSP)}, 2017, pp. 6533--6537.

\bibitem{HypergraphFreq}
S.~Zhang, Z.~Ding, and S.~Cui, ``Introducing hypergraph signal processing: {Theoretical} foundation and practical applications,'' \emph{IEEE Internet of Things Journal}, vol.~7, no.~1, pp. 639--660, 2020.

\bibitem{T-HGSP}
K.~Pena-Pena, D.~L. Lau, and G.~R. Arce, ``t-{HGSP}: Hypergraph signal processing using t-product tensor decompositions,'' \emph{IEEE Transactions on Signal and Information Processing over Networks}, vol.~9, pp. 329--345, 2023.

\bibitem{Karelia}
K.~Pena-Pena, L.~Taipe, F.~Wang, D.~L. Lau, and G.~R. Arce, ``Learning hypergraphs tensor representations from data via {t-HGSP},'' \emph{IEEE Transactions on Signal and Information Processing over Networks}, vol.~10, pp. 17--31, 2024.

\bibitem{Fuli}
F.~Wang, K.~Pena-Pena, W.~Qian, and G.~R. Arce, ``T-hyper{GNN}s: Hypergraph neural networks via tensor representations,'' \emph{IEEE Transactions on Neural Networks and Learning Systems}, vol.~36, no.~3, pp. 5044--5058, 2025.

\bibitem{HGNN+}
Y.~Gao, Y.~Feng, S.~Ji, and R.~Ji, ``{HGNN+}: General hypergraph neural networks,'' \emph{IEEE Transactions on Pattern Analysis and Machine Intelligence}, vol.~45, no.~3, pp. 3181--3199, 2023.

\bibitem{Volterra}
Q.~Yang, M.~Coutino, G.~Leus, and G.~B. Giannakis, ``Autoregressive graph volterra models and applications,'' \emph{EURASIP Journal on Advances in Signal Processing}, Jan 2023.

\bibitem{SPHINX}
\BIBentryALTinterwordspacing
I.~Duta and P.~Liò, ``{SPHINX}: Structural prediction using hypergraph inference network,'' 2024. [Online]. Available: \url{https://arxiv.org/abs/2410.03208}
\BIBentrySTDinterwordspacing

\bibitem{Bollengier}
M.~Bollengier, A.~A. Díaz~Berenguer, and H.~Sahli, ``Dynamic multi-hypergraph structure learning for disease diagnosis on multimodal data,'' in \emph{2024 46th Annual International Conference of the IEEE Engineering in Medicine and Biology Society (EMBC)}, 2024, pp. 1--5.

\bibitem{Zhang}
Z.~Zhang, H.~Lin, and Y.~Gao, ``Dynamic hypergraph structure learning,'' in \emph{Proceedings of the 27th International Joint Conference on Artificial Intelligence}, ser. IJCAI'18.\hskip 1em plus 0.5em minus 0.4em\relax AAAI Press, 2018, p. 3162–3169.

\bibitem{DualSmoothness}
B.~Tang, S.~Chen, and X.~Dong, ``Learning hypergraphs from signals with dual smoothness prior,'' in \emph{ICASSP 2023 - 2023 IEEE International Conference on Acoustics, Speech and Signal Processing (ICASSP)}, 2023, pp. 1--5.

\bibitem{Simplex1}
\BIBentryALTinterwordspacing
R.~Mulas, D.~Horak, and J.~Jost, \emph{Graphs, Simplicial Complexes and Hypergraphs: {Spectral} Theory and Topology}.\hskip 1em plus 0.5em minus 0.4em\relax Cham: Springer International Publishing, 2022, pp. 1--58. [Online]. Available: \url{https://doi.org/10.1007/978-3-030-91374-8_1}
\BIBentrySTDinterwordspacing

\bibitem{Simplex2}
S.~Gurugubelli and S.~P. Chepuri, ``Simplicial complex learning from edge flows via sparse clique sampling,'' in \emph{2024 32nd European Signal Processing Conference (EUSIPCO)}, 2024, pp. 2332--2336.

\bibitem{Simplex3}
A.~Buciulea, E.~Isufi, G.~Leus, and A.~G. Marques, ``Learning the topology of a simplicial complex using simplicial signals: {A} greedy approach,'' in \emph{2024 IEEE 13rd Sensor Array and Multichannel Signal Processing Workshop (SAM)}, 2024, pp. 1--5.

\bibitem{Simplex4}
------, ``Learning graphs and simplicial complexes from data,'' in \emph{ICASSP 2024 - 2024 IEEE International Conference on Acoustics, Speech and Signal Processing (ICASSP)}, 2024, pp. 9861--9865.

\bibitem{Tang}
B.~{Tang}, S.~{Chen}, and X.~{Dong}, ``Hypergraph structure inference from data under smoothness prior,'' \emph{arXiv e-prints}, p. arXiv:2308.14172, Aug. 2023.

\bibitem{SmoothnessTable}
C.~H. Nguyen and H.~Mamitsuka, ``Learning on hypergraphs with sparsity,'' \emph{IEEE Transactions on Pattern Analysis and Machine Intelligence}, vol.~43, no.~8, pp. 2710--2722, 2021.

\bibitem{GraphLP1}
A.~Elmoataz, O.~Lezoray, and S.~Bougleux, ``Nonlocal discrete regularization on weighted graphs: {A} framework for image and manifold processing,'' \emph{IEEE Transactions on Image Processing}, vol.~17, no.~7, pp. 1047--1060, 2008.

\bibitem{GraphLP2}
C.~Couprie, H.~Talbot, J.-C. Pesquet, L.~Najman, and L.~Grady, ``Dual constrained {TV}-based regularization,'' in \emph{2011 IEEE International Conference on Acoustics, Speech and Signal Processing (ICASSP)}, 2011, pp. 945--948.

\bibitem{StarCliqueSimilarity}
\BIBentryALTinterwordspacing
S.~Agarwal, K.~Branson, and S.~Belongie, ``Higher order learning with graphs,'' in \emph{Proceedings of the 23rd International Conference on Machine Learning}, ser. ICML '06.\hskip 1em plus 0.5em minus 0.4em\relax New York, NY, USA: Association for Computing Machinery, 2006, p. 17–24. [Online]. Available: \url{https://doi.org/10.1145/1143844.1143847}
\BIBentrySTDinterwordspacing

\bibitem{isomorphic}
\BIBentryALTinterwordspacing
K.~Hayashi, S.~G. Aksoy, C.~H. Park, and H.~Park, ``Hypergraph random walks, {Laplacians}, and clustering,'' in \emph{Proceedings of the 29th ACM International Conference on Information \& Knowledge Management}, ser. CIKM '20.\hskip 1em plus 0.5em minus 0.4em\relax New York, NY, USA: Association for Computing Machinery, 2020, p. 495–504. [Online]. Available: \url{https://doi.org/10.1145/3340531.3412034}
\BIBentrySTDinterwordspacing

\bibitem{Clique1}
\BIBentryALTinterwordspacing
Q.~Dai and Y.~Gao, \emph{Mathematical Foundations of Hypergraph}.\hskip 1em plus 0.5em minus 0.4em\relax Singapore: Springer Nature Singapore, 2023, pp. 19--40. [Online]. Available: \url{https://doi.org/10.1007/978-981-99-0185-2_2}
\BIBentrySTDinterwordspacing

\bibitem{Clique2}
S.~G. Aksoy, C.~Joslyn, C.~O. Marrero, B.~Praggastis, and E.~Purvine, ``Hypernetwork science via high-order hypergraph walks,'' \emph{EPJ Data Science}, vol.~9, no.~1, Jun 2020.

\bibitem{OneOfTheSmoothnessTerms}
M.~Hein, S.~Setzer, L.~Jost, and S.~S. Rangapuram, ``The total variation on hypergraphs - learning on hypergraphs revisited,'' in \emph{Proceedings of the 27th International Conference on Neural Information Processing Systems - Volume 2}, ser. NIPS'13.\hskip 1em plus 0.5em minus 0.4em\relax Red Hook, NY, USA: Curran Associates Inc., 2013, p. 2427–2435.

\bibitem{Duality}
N.~Komodakis and J.-C. Pesquet, ``Playing with duality: An overview of recent primal-dual approaches for solving large-scale optimization problems,'' \emph{IEEE Signal Processing Magazine}, vol.~32, no.~6, pp. 31--54, 2015.

\bibitem{CoraAndDBLP}
N.~Yadati, M.~Nimishakavi, P.~Yadav, V.~Nitin, A.~Louis, and P.~Talukdar, \emph{HyperGCN: a new method of training graph convolutional networks on hypergraphs}.\hskip 1em plus 0.5em minus 0.4em\relax Red Hook, NY, USA: Curran Associates Inc., 2019.

\bibitem{Lung}
\BIBentryALTinterwordspacing
``Lung cancer trends brief: Mortality,'' American Lung Association. [Online]. Available: \url{https://www.lung.org/research/trends-in-lung-disease/lung-cancer-trends-brief/lung-cancer-mortality-(1)}
\BIBentrySTDinterwordspacing

\bibitem{Smoking}
\BIBentryALTinterwordspacing
``State tobacco activities tracking and evaluation (state) system,'' Centers for Disease Control and Prevention. [Online]. Available: \url{https://nccd.cdc.gov/STATESystem/}
\BIBentrySTDinterwordspacing

\bibitem{Temperature}
\BIBentryALTinterwordspacing
``Statewide mapping | climate at a glance,'' National Centers for Environmental Information. [Online]. Available: \url{https://www.ncei.noaa.gov/access/monitoring/climate-at-a-glance/statewide/mapping}
\BIBentrySTDinterwordspacing

\bibitem{LungSmokeCorrelation}
\BIBentryALTinterwordspacing
``Lung cancer risk factors,'' Centers for Disease Control and Prevention, 10 2024. [Online]. Available: \url{https://www.cdc.gov/lung-cancer/risk-factors/index.html}
\BIBentrySTDinterwordspacing

\bibitem{Purdue}
\BIBentryALTinterwordspacing
M.~F. Baumgardner, L.~L. Biehl, and D.~A. Landgrebe, ``220 band {AVIRIS} hyperspectral image data set: June 12, 1992 indian pine test site 3,'' Sep 2015. [Online]. Available: \url{https://purr.purdue.edu/publications/1947/1}
\BIBentrySTDinterwordspacing

\bibitem{HyperspectralData}
\BIBentryALTinterwordspacing
``Hyperspectral remote sensing scenes,'' www.ehu.eus. [Online]. Available: \url{https://www.ehu.eus/ccwintco/index.php?title=Hyperspectral_Remote_Sensing_Scenes#Indian_Pines}
\BIBentrySTDinterwordspacing

\bibitem{HyperspectralMethod}
Z.~Ma, Z.~Jiang, and H.~Zhang, ``Hyperspectral image classification using feature fusion hypergraph convolution neural network,'' \emph{IEEE Transactions on Geoscience and Remote Sensing}, vol.~60, pp. 1--14, 2022.

\bibitem{SLIC1}
L.~Ma, Q.~Wang, J.~Zhang, and Y.~Wang, ``Parallel graph attention network model based on pixel and superpixel feature fusion for hyperspectral image classification,'' in \emph{IGARSS 2023 - 2023 IEEE International Geoscience and Remote Sensing Symposium}, 2023, pp. 7226--7229.

\bibitem{SLIC2}
X.~Sun, F.~Zhang, L.~Yang, B.~Zhang, and L.~Gao, ``A hyperspectral image spectral unmixing method integrating {SLIC} superpixel segmentation,'' in \emph{2015 7th Workshop on Hyperspectral Image and Signal Processing: Evolution in Remote Sensing (WHISPERS)}, 2015, pp. 1--4.

\bibitem{GCNHSI}
D.~Hong, L.~Gao, J.~Yao, B.~Zhang, A.~Plaza, and J.~Chanussot, ``Graph convolutional networks for hyperspectral image classification,'' \emph{IEEE Transactions on Geoscience and Remote Sensing}, vol.~59, no.~7, pp. 5966--5978, 2021.

\end{thebibliography}
